\documentclass[11pt]{article}

\usepackage[margin=1.1in]{geometry}
\usepackage[T1]{fontenc}
\usepackage[utf8]{inputenc}
\usepackage{newtxtext,newtxmath}
\usepackage{microtype}
\usepackage{graphicx}
\usepackage{booktabs}
\usepackage{array}
\usepackage{amsmath}
\usepackage{xcolor}
\usepackage[numbers,sort&compress]{natbib}
\usepackage[colorlinks=true,linkcolor=blue!60!black,citecolor=blue!60!black,urlcolor=blue!60!black]{hyperref}
\usepackage{caption}

\usepackage{listings}
\usepackage{enumitem}
\setlist{itemsep=2pt,topsep=4pt}

\newcommand{\mimeo}{\textsc{mimeo}}

\title{\mimeo: Compiling Public Expert Corpora into Agent Skills\\and Testing What Transfers}

\author{%
  Timothy Kassis\\
  K-Dense, Inc.\\
  \texttt{timothy.kassis@k-dense.ai}
}

\date{}

\begin{document}
\maketitle

% Graphical abstract. Deliberately not a float and not numbered: it sits above
% the abstract on page 1, so the figure counter still starts in the body. The
% vertical spacing is tuned to keep the whole front matter on one page.
\vspace{-1.5\baselineskip}
\begin{center}
  \includegraphics[width=\textwidth]{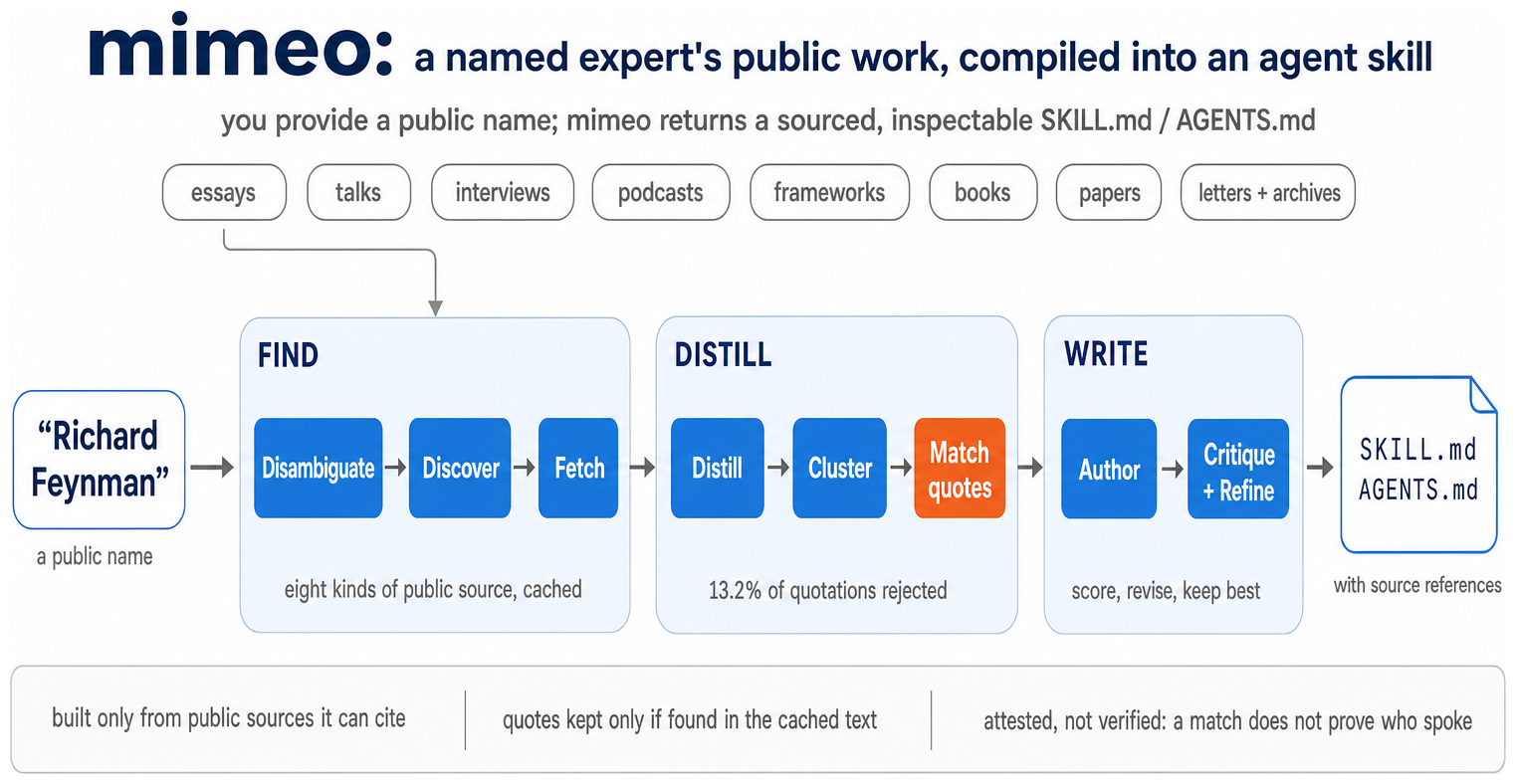}
\end{center}
\vspace{-0.6\baselineskip}

\begin{abstract}
Giving an agent a file about a named expert can supply hard-to-find material,
produce a recognizable persona, or change what the agent decides. These are
different claims. We test each one.
\mimeo{} is an open-source tool that finds a person's public work, checks each
extracted quotation against the cached source text, and writes a file an agent
can load.
Eight logged builds averaged 38 model calls; the check rejects 13.2\% of
extracted quotations. We tested four expert files with one coding-agent
harness. Knowledge access was clearest: \mimeo{} answered all 20 obscure,
quotation-heavy questions; no closed-book condition answered more than 10.
Keyword search (BM25) over the same pages answered 15--17, a gap this sample
cannot resolve. Grounding showed one clear benefit: personas written from model
memory misstated a documented position on
1--4 of 20 answers under every grader; the plain agent and \mimeo{} never did.
Every persona was easy to spot on short open prompts, and adding task material
lowered identification by 18--23 points. \mimeo{} was no more
identifiable than a from-memory profile. Judgment transfer remained unresolved
because both tests hit their ceiling: every condition found 94--97\% of the
problems planted in engineering tasks and scored 94--100\% on 16 new
application scenarios. An AI-judged ``sounds like the expert'' score
changed with the judge: two of four preferred answers based on a model's
stereotype, while two found no difference on the same text. That is a caution
against relying on a single AI judge.
The evidence supports \mimeo{} as a compact, inspectable reference
on a person, not as a demonstrated transfer of their judgment. Toolkit and
expert profiles:
\url{https://github.com/K-Dense-AI/mimeo}
\end{abstract}

\clearpage

\section{Introduction}
\label{sec:intro}

Suppose a small AI lab asks an agent to choose its next research direction. The
lab first gives the agent a multi-page file compiled from Richard Sutton's
public work. That file might help the agent quote Sutton, make its answer
recognizably Sutton-like, or change the research direction it recommends.
These are three different outcomes: \emph{knowledge access}, a
\emph{recognizable persona}, and \emph{judgment transfer}.

The larger question is how to encode an expert's knowledge so an agent can use
it while working. A public record captures what the expert chose to write or
say, not everything that decades of practice taught them.
Some expertise is tacit: people can use it without being able to state it
fully~\citep{polanyi2009tacit}. Expertise includes facts and procedures as well
as selection: what to notice, which problem is worth
pursuing, how to approach it, and when to stop. We use \emph{taste} for this
selection. A capable base model may have many possible moves; an expert skill
could help it choose among them. Whether a text file can do that is the
judgment-transfer question in this paper.

Agent products increasingly read project instructions and reusable skills from
markdown files such as \texttt{AGENTS.md} and \texttt{SKILL.md}
\citep{anthropic2025skills, agentsmd2025}. We present \textbf{\mimeo}, an
open-source tool that turns a name and public web sources into either kind of
file. It identifies the person, searches eight kinds of material, groups
recurring ideas, checks longer quotations against cached source text, and
revises the draft against an editorial checklist. The output carries source
references, but it is not a verified account of a person. Cached text may be
only an excerpt, and a text match proves neither who spoke nor whether the
surrounding claim follows. The released gallery also contains source
identifiers that point nowhere. In one unreleased re-run, the quotation check
ran and passed a quotation that the file still credited to the wrong source.

We test the three outcomes using one coding-agent harness, four original expert
profiles, and four LLM judges (Section~\ref{sec:results}). Corpus access clearly
improves recall of obscure wording over answering from memory, but a simple
BM25 keyword search recovers most of that gain. Grounding shows one distinct
benefit: asked about obscure documented positions, both personas written
without sources name a position the record does not document, while the plain
agent and \mimeo{} decline. A recognizable persona appears
on short open prompts whether the profile comes from sources or model memory;
task material makes every persona harder to spot. Judgment transfer remains
unresolved because both tests of it hit their ceiling. The measure itself also
depends on the judge: two judges prefer stereotype-based answers on the pooled
``sounds like the expert'' score, while two detect no difference on the same
text. These results support \mimeo{} as a compact, inspectable reference on a
person. They do not show that it transfers the person's judgment or works
better than looking up the source material when needed.

\paragraph{Related work.}
Hand-designed role prompts sometimes help and often do
not~\citep{kong2024roleplay, xu2023expertprompting, zheng2024helpful,
hu2024quantifying, persona2026retrieval}. Paired evaluations of agent skills
also conflict: one reports a large gain on tasks with automatic
verifiers~\citep{skillsbench2026}, while others find almost no effect on real
software-engineering instances~\citep{sweskillsbench2026,
eval2026agentsmd}. \mimeo{} asks a different question: what transfers when the
file is a sourced, multi-page profile rather than a role label or a procedure
optimized against a task score? Appendix~\ref{sec:related} gives the full
discussion.

\section{The \mimeo{} System}
\label{sec:system-summary}

\mimeo{} is a Python command-line tool. The command
\texttt{mimeo "Richard Feynman"} creates a ready-to-install skill directory in
eight cached stages. The tool first identifies the person, then searches essays,
talks, interviews, podcasts, frameworks, books, papers, and archives. It
fetches whatever text is available, turns each record into labeled ideas and
quotations, groups recurring ideas across records, and writes
\texttt{SKILL.md}, \texttt{AGENTS.md}, or both. A final editing loop revises
the file against a checklist. Every stage is cached, so an interrupted or
changed run repeats only the work whose inputs changed.
Appendix~\ref{sec:system} describes the full pipeline.

The output is prose under headed sections, which an agent reads as it would any
other instruction file. These are the opening lines of the generated file for
Andrej Karpathy:

\noindent
\begin{lstlisting}
# Thinking like Andrej Karpathy

## Core principles
* Build from Scratch to Understand: To truly grasp complex
  systems, you must manually implement the core algorithms
  without relying on automated tools [...]

## Anti-patterns they push against
* Jumping to Full Autonomy: Trusting an AI to generate massive,
  unverified outputs (like a 10,000-line code diff) creates a
  massive verification bottleneck for the human.
\end{lstlisting}

\noindent
Appendix~\ref{app:example} gives a longer excerpt with its source references.

The quotation stage checks each quoted passage against the cached text of its
listed records and drops passages that do not match. A match makes a quotation
\emph{attested}, not \emph{verified}: it does not prove who spoke, whether the
claim around the quotation follows, or whether the record belongs to the right
person~\citep{liu2023verifiability, gao2023alce}. Experiments E1 and E3 test
these safeguards. The eight logged re-runs average 37.8 LLM calls and 7.1
minutes, and quotation matching removes 41 of 311 extracted spans (13.2\%).
With matching disabled, the output contains two invented sentences among 36
quotations, both attributed to living people. With matching enabled, none of
34 quotations is invented; one matched quotation is still credited to the
wrong source. The checks catch real failures, but they do not turn the output
into a verified account. Appendices~\ref{sec:intrinsic} and
\ref{sec:extrinsic-ablations} give the full results.

\section{Results}
\label{sec:results}

\begin{figure}[!ht]
  \centering
  \includegraphics[width=\textwidth]{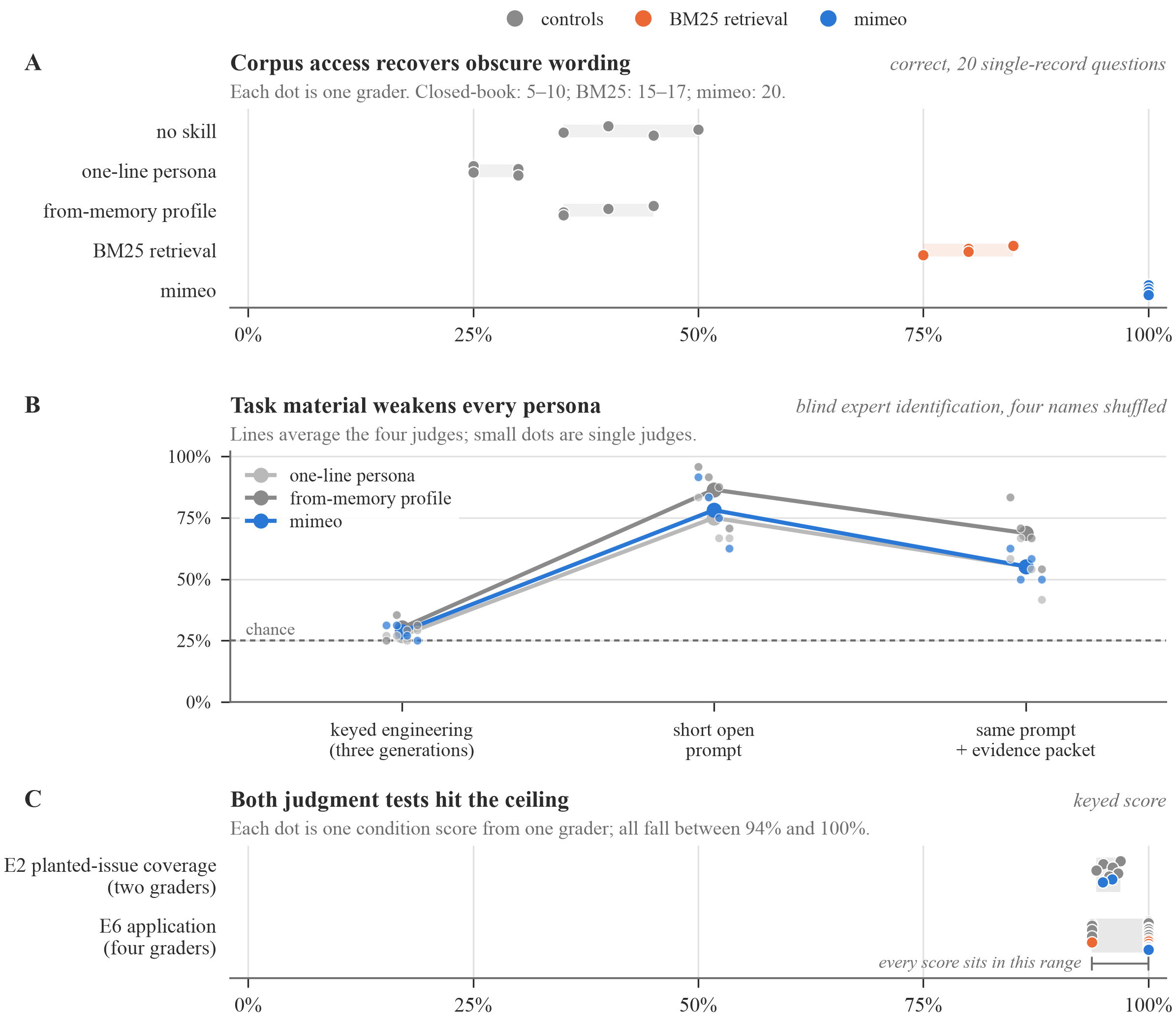}
  \caption{\textbf{What transfers depends on what is measured.}
  Gray marks the controls (no skill, one-line persona, from-memory profile);
  orange marks BM25 retrieval; blue marks \mimeo{}. (a)~Each point is one
  grader's score on the same 20 questions. (b)~Blind identification of the
  loaded expert across three task settings: the repeated keyed engineering
  tasks, the short open prompts, and the same prompts with an evidence packet
  attached. Lines average the four judges, small dots are single judges, and
  the dashed line is the 25\% chance rate. (c)~Each point is one
  condition under one grader. Repeated grades are shown separately, not
  treated as independent data.}
  \label{fig:headline}
\end{figure}

Six studies test the three claims separately, with three smaller arms probing
familiarity, task context, and deployment. E1/E3 examine the artifact itself;
the rest test behavior. Table~\ref{tab:studies} groups every study under the
claim it speaks to. Except for E4b's familiarity
follow-up, the behavioral studies use one agent (\texttt{claude-sonnet-5}) and
four original experts. The four main
conditions are \textbf{no skill}, \textbf{one-line persona},
\textbf{from-memory profile}, and \textbf{\mimeo}. Four models score the
outcomes. We call them graders where a written key fixes the right answer (E4,
E6, and, with two of the four, E2 coverage) and judges where the score is their
own comparison (identification and pairwise quality). We treat a finding as
established only if every assigned model agrees.
Figure~\ref{fig:headline} summarizes the main results; the appendices give each
design, estimate, and protocol in full.

\begin{table}[!ht]
  \centering\small
  \setlength{\tabcolsep}{4pt}
  \renewcommand{\arraystretch}{1.12}
  \begin{tabular}{@{}l>{\raggedright\arraybackslash}p{3.6cm}>{\raggedright\arraybackslash}p{3.3cm}>{\raggedright\arraybackslash}p{5.6cm}@{}}
    \toprule
    Study & What it asks & Scale & Outcome in one line \\
    \midrule
    \multicolumn{4}{@{}l}{\emph{The artifact and its safeguards}} \\
    E1 & What does the pipeline produce? & 22-expert gallery; 8 logged builds &
    The quotation check removes 13.2\% of extracted spans; the revision loop
    raises only its own checklist score. \\
    E3 & What breaks when pipeline stages are removed? & 3 experts rebuilt &
    Without matching, invented quotations ship; clustering is invisible to the
    checklist. \\
    E2 on-demand & Does the agent open the skill on its own? & 16 runs & It
    loads the skill in 5 of 16 runs. \\
    \addlinespace
    \multicolumn{4}{@{}l}{\emph{Knowledge access}} \\
    E4 & Can the agent recall obscure documented positions? & 40 questions, 5
    conditions, 4 graders & \mimeo{} answers all 20 single-record questions;
    BM25 15--17; closed-book at most 10. \\
    E4b & Does the recall gap survive less famous experts? & 3 experts, 18
    questions, 2 graders & The gap holds; the arm is exploratory. \\
    \addlinespace
    \multicolumn{4}{@{}l}{\emph{Recognizable persona}} \\
    E5 & Is the persona recognizable on open prompts? & 6 prompts, 4 judges &
    Every persona condition is identified well above chance. \\
    E5b & Does task material weaken the persona? & same prompts, plus
    6.9--7.1k characters of evidence & Identification drops 18--23 points;
    every judge records a drop. \\
    \addlinespace
    \multicolumn{4}{@{}l}{\emph{Judgment transfer}} \\
    E2 & Does an installed profile change engineering work? & 8 tasks $\times$
    4 experts; keyed tasks run three times & Coverage is 94--97\% in every
    condition, at the ceiling; pairwise quality is unresolved. \\
    E6 & Do corpus principles change new decisions? & 16 scenarios, 5
    conditions, 4 graders & 94--100\% in every condition, at the ceiling;
    transfer unmeasured. \\
    \bottomrule
  \end{tabular}
  \caption{The studies at a glance, grouped by the claim each one speaks to.
  E1 and E3 examine the artifact and its safeguards; the rest test behavior in
  one coding-agent harness with four experts. The appendices give each design,
  estimate, and protocol in full.}
  \label{tab:studies}
\end{table}

\subsection{Knowledge access: the corpus helps, but distillation is unresolved}

The artifact includes inspectable source references. E1/E3 show why its
quotation check matters (Section~\ref{sec:system-summary}). Under \mimeo{}, the
wording of 35 of 38 E4 quotations appears in the cached corpus. Runtime
retrieval gives 37 of 43, while the closed-book conditions range from
28--50\%. A match makes a quotation attested, not verified
(Section~\ref{sec:system-summary}).

E4 asks 20 widely repeated questions and 20 questions drawn from one record.
All four graders give \mimeo{} 20/20 on the single-record set, while no
closed-book condition exceeds 10/20. Each question was written from an item in
the same corpus the \mimeo{} file was built from, so a ceiling score for
\mimeo{} follows from the design rather than testing it
(Appendix~\ref{sec:regimes-knowledge}). What the questions do test is every
condition that was not built from the key. Corpus access therefore improves
obscure recall over answering from memory. It does not yet establish a benefit
from distillation. Given only the question, a plain BM25 search over the same
cached records answers 15--17. \mimeo{} leads by three to five questions and
loses none, but exact paired tests do not separate them. The sample is too small
to tell whether the static profile beats looking up passages when needed.

The closed-book conditions fail differently. On these 20 questions, the agent
with no skill never attributes a position other than the documented one, under
all four graders, and neither does \mimeo{}; the one-line persona does so on
1--3 answers and the from-memory profile on 3--4, under every grader. An
ungrounded persona file moves the agent from declining to naming the wrong
position. The graders worked from a positive key, so this count includes an
answer that offers a different plausible position as well as one that
contradicts the key (Appendix~\ref{sec:regimes-knowledge}). Zero against three
of 20 probes is not significant on its own. What supports the claim is that all
four graders record the same pattern in both persona conditions. This is the one
measure where grounding the profile in sources clearly beats writing it from the
model's memory.

\subsection{Recognizable persona: task material weakens every persona}

E5 asks six short advisory questions with no attached material. Every persona
condition is identifiable above the 25\% chance rate, whether it is a one-line
prompt, a from-memory profile, or \mimeo{}. Basing the profile on sources adds
no visible advantage over the from-memory profile.

E5b keeps the requests unchanged and adds 6.9--7.1k characters of task
evidence. Identification then drops by 20 points for the one-line persona, 18
for the from-memory profile, and 23 for \mimeo{}, averaged across judges. Every
judge records a drop for every persona. Flipping the signs of the six task
differences in every possible way gives an exact $p=.031$, while every point
estimate remains above chance. The persona survives, but task material competes
with it.

Identification is weaker on E2's richer engineering tasks. Across three
generations, judges identify \mimeo{} in 25--31\% of answers. Every interval
formed by resampling task means includes the 25\% chance rate. The measure is a
forced choice among the four shuffled names, so a judge cannot decline. The
result is a weak, unresolved persona signal in answers shaped by the task
materials.

\subsection{Judgment transfer: both tests hit their ceiling}

In E2, four engineering tasks have fixed lists of planted issues. Across three
generations and two graders, every condition finds 94--97\% of the list.
Equivalence tests rule out a ten-point coverage difference between \mimeo{} and
each baseline on these tasks. Yet coverage under no skill is already at
95--97\%, so only three to five points remain for improvement. E2 rules out a
large coverage loss; it cannot show whether the profile changes harder
decisions.

E6 removes the quotation demand and asks 16 new questions that require applying
corpus principles. Again, every condition, including no skill and BM25
retrieval, scores 94--100\% under all four graders. Under no skill, the agent
already gives the keyed recommendation almost every time. E6 therefore gives no
evidence either for or against judgment transfer. Pairwise advice quality and
``sounds like the expert'' judgments also do not resolve the question:
Appendix~\ref{app:stats} states the statistical procedures, and
Appendix~\ref{app:discussion} shows that those results depend on which judge is
asked.

\section{Discussion}
\label{sec:discussion}

\paragraph{What transfers?}
Corpus access helps: \mimeo{} supplies obscure wording that the model often
cannot recall. What remains unresolved is whether a static distilled file
works better than looking up the same material when needed. Persona recognition
is a separate result. A persona can be recognizable without being unique to a
sourced profile. Short prompts expose every persona, the from-memory profile is
at least as identifiable as \mimeo{}, and task material weakens all three.
Judgment transfer remains open. The engineering and application suites both
hit their ceiling. They rule out a large E2 coverage loss on these tasks but
leave little room to observe a gain.

A profile can give an agent useful material without making it decide as the
expert would. It can also change an answer's voice without improving its
reasoning. On engineering
tasks, the weaker persona may reflect competition from task materials, a poor
match between expert and task, or advice the base model already knows. The
evidence covers four profiles, four keyed tasks, and one commercial agent model
as it was served during the study; it is not a universal null.

\paragraph{Taste is part of judgment transfer.}
Model capability and expert taste are separate. A base model may know many
available methods without sharing an expert's taste. Taste is the expert's
selection among them: what deserves attention, which trade-off matters, when a
standard procedure does not fit, and when there is enough evidence to act. In
science, this includes choosing promising problems. Recent work captures one
part of scientific taste as a preference for ideas with potential long-term
impact, then learns that preference from community feedback~\citep{tong2026taste}.
A profile skill takes another route: it puts one
expert's criteria into the model's context at inference time. Our experiments
do not show that this works. In our framework, injecting taste is judgment
transfer, not knowledge access or a recognizable voice.

\begin{table}[t]
  \centering\small
  \setlength{\tabcolsep}{4pt}
  \begin{tabular}{llcccc}
    \toprule
    Measure & Suite & \texttt{opus-5} & \texttt{gpt-5.6} & \texttt{grok-4.6} & \texttt{ds-v4} \\
    \midrule
    Sounds like the expert & E2+E5 & .37 & .51 & .41 & .47 \\
    Advice quality vs.\ no skill & E5 & .23 & .17 & .50 & .61 \\
    Single-record recall, \mimeo{} & E4 & 20/20 & 20/20 & 20/20 & 20/20 \\
    Single-record recall, BM25 & E4 & 17/20 & 16/20 & 16/20 & 15/20 \\
    Forced ID, \mimeo{} (3 generations) & E2 & .31 & .31 & .27 & .25 \\
    Forced ID, \mimeo{} & E5 & .92 & .83 & .75 & .63 \\
    Forced ID, \mimeo{} + packet & E5b & .63 & .50 & .58 & .50 \\
    \bottomrule
  \end{tabular}
  \caption{The four judges, fixed in advance. Pairwise rows report \mimeo{}'s
  share of the vote, recall rows count correct answers, and identification rows
  give accuracy when the four names are shuffled (chance .25). Measures with an
  outside key keep the same ordering across judges; advice quality and
  ``sounds like the expert'', which have no key, depend on who is asked.}
  \label{tab:judges}
\end{table}

\paragraph{Whether an answer sounds like the expert depends on who is asked.}
Four judges score the same 112 answer pairs differently
(Table~\ref{tab:judges}). Two significantly
prefer stereotype-based answers on the pooled ``sounds like the expert''
measure; two find no
difference, and none prefers \mimeo{}. The spread between judges is larger than
we see after shuffling judge labels within pairs ($p=9.5\times10^{-4}$).
Measures with an outside check, such as a fixed issue list, answer key, source
text, or shuffled expert names, are more stable. ``Sounds like the expert'' has no such
key. A panel can reveal that instability but cannot tell us what Karpathy or
Sutton would actually do. Appendix~\ref{app:discussion} reports the agreement
measures in full.

\paragraph{Evaluating persona files.}
A test of an expert-persona file should include a from-memory profile written
by the same model, so it compares grounding with the stereotype it competes
with. A retrieval control over the same corpus compares a static file with
looking things up. Tasks should
leave room for improvement, or a ceiling will hide any effect. Identification
should be a forced choice among shuffled names, since order and abstention
otherwise drive the score. Judged measures with no outside key need more than
one judge, and the panel should span multiple developers. A claim should
survive the whole panel before it is reported as a finding.

\paragraph{Beyond public records.}
\mimeo{} can distill only what has entered the public record. Decades of
scientific or medical practice may also produce situated knowledge that never
appears in a paper, talk, or interview. Cognitive task analysis uses structured
interviews about specific events and methods to represent the cues and
decisions behind proficient work~\citep{hoffman1998critical,
crandall2006working}. A later system could combine \mimeo{}'s source trail with
consented case walkthroughs, repeated interviews, and months of observation as
an agent shadows an expert at work. It could ask about decisions while their
context is still available, then turn recurring cues, exceptions, and
trade-offs into a skill the expert can inspect and correct. In medicine, this
would require strict consent, privacy, and safety controls. Recording a
rationale would not make it safe to delegate clinical judgment.

\paragraph{Next experiments.}
The next public-corpus test should study decisions where the expert's
documented view departs from the model's default, using raters who know the
expert's work. A larger retrieval study should hold the context budget fixed
and compare the static file with keyword, semantic-vector, and combined
retrieval. Rebuilding each artifact several times would show how much the
results depend on which sources the pipeline finds and how it writes the
profile. A test of an interviewed or shadowed expert should instead hold out
later cases and compare the expert's and agent's choices, including which cues
they notice and when they stop. A recognizable voice would not validate that
kind of transfer.

\section{Ethics and Limitations}
\label{sec:ethics-summary}

\mimeo{} builds artifacts about identifiable, mostly living people. Public
material is not consent for an AI-written profile. A person's web record is
partial, dated, and shaped by which languages and venues can be searched.
Distilling that record can freeze a position the person has since abandoned or
turn a conditional remark into a rule. Our probes show one result of this risk:
misstating the position a record documents. Grounding reduces that error. Asked
about obscure documented positions, the two
personas written without sources misstate one on 1--4 of 20 answers under every grader, while the plain agent and \mimeo{} do so zero
times. The named person did not write, approve, or endorse the artifact.
Neither the toolkit nor the gallery lets them review, correct, or remove one.
This is an unresolved defect in how the system is deployed, not polish to add
later. Quotation matching reduces one risk but does not remove it: closely
matching text does not establish who spoke, and the system has already credited
a quotation to the wrong source. Fetched pages are untrusted input. The markers
the pipeline puts around them are not a security boundary. Users remain
responsible for source licenses and takedowns.

The study covers one commercial agent model that keeps changing, one harness,
four experts, four keyed tasks, and one built artifact per expert. The paper's
author wrote the tasks and keys. Repeated answers within one task are not
independent~\citep{baayen2008mixed}, so we summarize contrasts at the task
level. The coverage and application tests both sit at their ceiling. E4 mainly
tests quotation recall drawn from the same corpus the treatment carries. No
rater in this study is an expert on all four people, so we do not call the
artifacts digital twins, clones, or verified expert reasoning.
Appendix~\ref{sec:ethics} gives the full treatment.

\section{Conclusion}
\label{sec:conclusion}

\mimeo{} turns public records about a named expert into a portable,
source-carrying \texttt{SKILL.md} or \texttt{AGENTS.md}. The result is an
inspectable reference, not a verified copy of how the person reasons. Web
records may be partial, a matched quotation is attested rather than verified
(Section~\ref{sec:system-summary}), and the generated gallery has known
attribution failures.

Each claim ends somewhere different. Corpus access helps, but a keyword search
over the same records recovers most of the gain, so the value of distillation
itself is unresolved. The persona is recognizable and not unique to \mimeo{}:
a profile written from the model's memory is at least as identifiable, and task
material weakens all three. On repeated engineering tasks, identification
falls to 25--31\%, which these data cannot separate from the 25\% chance rate.
Judgment transfer is not demonstrated,
because both tests of it sit at their ceiling and the judgment-based
``sounds like the expert'' measure changes with the judge. Grounding wins
clearly on one measure only: the ungrounded personas assert positions the
record does not document, where the plain agent and \mimeo{} decline.

\mimeo{} addresses one part of a larger knowledge-encoding problem: what an
expert made public. Capturing tacit expertise and taste would require
interaction with the expert's consent, evidence from decisions in context, and tests of
whether the agent makes the same choices. The present results establish none of
that. They separate these questions from recall and persona so later systems
can test them directly.

These conclusions draw on 732 agent runs. Pairwise and identification
judgments, along with the main E4 and E6 grades, were repeated across four
judges; E2 coverage and E4b were graded twice. The toolkit and the expert
profiles it produces are open source at
\url{https://github.com/K-Dense-AI/mimeo}.

\bibliographystyle{unsrtnat}
\bibliography{refs}

\appendix
\section{Related Work}
\label{sec:related}

\paragraph{Persona and role prompting.}
Hand-written role-play prompts can elicit useful step-by-step reasoning and
improve performance without examples~\citep{kong2024roleplay}.
ExpertPrompting also found that GPT-4 preferred answers produced under
LLM-written ``distinguished expert'' identities to vanilla
answers~\citep{xu2023expertprompting}. The wider
evidence is mixed. In a study of 162 personas and four model families, adding a
persona line to the system prompt did not reliably improve accuracy. Most
effects were indistinguishable from noise~\citep{zheng2024helpful}. Details
about a simulated person's identity also explain little of the variation in
the human judgments being simulated~\citep{hu2024quantifying}. Studies that
separate different kinds of help likewise find mixed
results~\citep{luzdearaujo2024helpful, luzdearaujo2025principled}. Across nine
models and 27 tasks, expert personas usually help or have no significant
effect. Yet models respond strongly to irrelevant persona details and do not
consistently reflect the relevant details in their
answers~\citep{luzdearaujo2025principled}.

A more recent analysis covers 1{,}140 open-ended questions and 38 expert roles.
It finds that role prompting reshapes an answer rather than improving it: the
answer reads as more expert but becomes less clear. The effect depends on the
setting. Role prompting works best for advisory questions and in fields such as
medicine and psychology, while plain prompting wins for technology, science,
finance, and legal questions~\citep{persona2026retrieval}. Software engineering
is closest to that study's broad technology category, where the baseline
advantage appears on conceptual and explanatory questions. The comparison is
therefore suggestive but indirect. Other work finds that expert personas can
improve judged alignment while reducing accuracy~\citep{prism2026}. In our E5
data, two judges find a quality penalty for \mimeo{} relative to no skill. The
other two do not, so we do not count the penalty as replicated.

Most of this work studies short, ungrounded personas consisting of a name and a
role label. \mimeo{} instead tests a multi-page profile built from attributed
public records, with quotations checked against the source text. The question
is whether this profile does anything that a short, ungrounded persona does not.
Our evaluation covers the advisory setting, where the literature finds that
persona effects are possible, and includes the one-line persona as an explicit
baseline.

Another line of work changes a model's character through its internal
activations instead of through text in its context. Representation
engineering~\citep{zou2023repe} and contrastive activation
addition~\citep{rimsky2024caa} push behavior along directions found inside the
model. Persona vectors extract one such direction from a trait name, then use it
to monitor and control drift in character~\citep{chen2025personavectors}. We
study a different intervention, a \emph{file}, to test whether writing down how
an expert thinks and loading that description as context transfers judgment.
Our limited results do not address whether changing weights or activations can
control a persona.

\paragraph{Agent skills and repository context files.}
Agent Skills are \texttt{SKILL.md} files with YAML front matter, where the agent
sees a short description first and loads the body only when it decides the skill
is relevant. Anthropic introduced them in October 2025 and published them as an
open standard that December; more than forty agent products now support the
format~\citep{anthropic2025skills}. The complementary \texttt{AGENTS.md} convention
has no schema and is always loaded. Unlike Agent Skills, it is not one vendor's
product: it emerged across several coding-agent teams and is now overseen by
the Agentic AI Foundation~\citep{agentsmd2025}. Research on these files remains
limited. One study found that over 99\% of the \texttt{SKILL.md} files it
surveyed had at least one authoring defect described as a ``skill
smell''~\citep{skillsmells2026}. Some of those defects overlap with the checks
in \mimeo{}'s critique stage. A controlled evaluation using tasks from
SWE-bench~\citep{jimenez2024swebench} found that generic repository-level
context files did not generally improve outcomes. They also raised the cost of
running the model by over 20\%~\citep{eval2026agentsmd}. We use the same paired
on/off design for a different kind of file: a profile of how an expert reasons
rather than a repository overview. We also study judgment tasks rather than
issue resolution.

Paired on/off evaluations of skills now report conflicting results. The current
SkillsBench report pairs 87 tasks that have automatic verifiers with curated
skills.
It reports a large average gain, from a 33.9\% pass rate to
50.5\%~\citep{skillsbench2026}. SWE-Skills-Bench uses the same design with 49
public skills and roughly 565 real software-engineering instances, but finds
almost no effect. Of the 49 skills, 39 produce no improvement at all. The mean
gain is $+1.2\%$, while token overhead reaches
451\%~\citep{sweskillsbench2026}. Another study traces outcomes to individual
skills across both benchmarks. It attributes 307 failures and cost regressions
to the loaded skill and finds that the damaging skills are usually topically
relevant~\citep{skillharm2026}. The benchmark with checkable tasks shows gains;
the real-software benchmark often does not. These studies differ in more than
the kind of guidance they provide, so they do not show whether guidance type
explains the contrast. Our study sits at the judgment-focused end: it examines
a description of how a person thinks. Its
paired E2, E5/E5b, and E6 evaluations use advisory tasks, not issue resolution.
Position papers and early surveys describe skills as a way to provide reusable
knowledge to agents~\citep{knowledgeactivation2026, skillarch2026,
skillsurvey2026, skillevalsurvey2026}; \mimeo{} applies that idea to individual
human experts.

\paragraph{Automatic prompt engineering.}
APE~\citep{zhou2022ape}, OPRO~\citep{yang2023opro},
PromptBreeder~\citep{fernando2023promptbreeder}, DSPy~\citep{khattab2023dspy},
and GEPA~\citep{agrawal2025gepa} tune prompts against a task metric using
search, evolution, or compilation. \mimeo{} addresses a different problem. It
distills a body of work into a prompt without using any task metric. The goal is
faithfulness to the corpus rather than a benchmark score. Its critique loop is
closer to rubric-scored Self-Refine~\citep{madaan2023selfrefine} than to
metric-driven search. The approaches could be combined by using a distilled
skill to seed a metric-driven optimizer.

\paragraph{Skills learned from agent experience.}
Voyager grows a skill library from exploration in an embodied
world~\citep{wang2023voyager}. ExpeL distills lessons from the agent's own
past runs~\citep{zhao2023expel}, and Agent Workflow Memory turns records of
execution into reusable workflows~\citep{wang2024awm}. Systems released in
2026, including CODESKILL~\citep{codeskill2026}, evolve skills for coding agents
from their own experience. SkillGen builds an auditable skill from a base
agent's successful and failed runs. It validates the skill as an intervention
by comparing the same instances with and without it, so both repairs and
regressions count~\citep{skillgen2026}. We adopt that paired design in
Appendix~\ref{sec:extrinsic}. SkillGen applies it to skills generated from an
agent's own experience; we apply it to profiles built from another person's
public record. SkillGenBench makes it possible to benchmark skill-generation
pipelines themselves~\citep{skillgenbench2026}. The intervention design is the
same even though the source material differs.

\paragraph{Context as a way to supply knowledge.}
Other work uses context to supply knowledge rather than steer behavior. LLMs
learn rare facts less reliably, and their accuracy tracks how often a fact
appears in pretraining~\citep{kandpal2023longtail, sun2024headtotail}.
Retrieval-augmented generation bases an answer on records fetched when the
query arrives~\citep{lewis2020rag}. Supplying knowledge in context can also
outperform unsupervised fine-tuning: in one comparison, retrieval did so for
both existing and new information~\citep{ovadia2023finetuning}. Models often,
but not always, follow evidence in context over what they
memorized~\citep{xie2024chameleon}. \mimeo{} performs the retrieval and
compression offline, then installs a static file.
Appendix~\ref{sec:regimes-knowledge} compares that file with BM25 over the same
cached records, using only the question. The simple retriever recovers most of
the direct-recall gain. Our evidence therefore supports access to the corpus
more strongly than the editing performed on that corpus.

Matching quotations addresses only one part of grounding an answer in sources.
GopherCite retrieves supporting quotations~\citep{menick2022verified}; RARR
edits claims after the fact against retrieved evidence~\citep{gao2023rarr}; ALCE
scores whether citations support their statements and whether every statement
is supported~\citep{gao2023alce}; and FActScore splits long text into small
claims so that each one can be checked against a
source~\citep{min2023factscore}. Our matcher is narrower. It asks whether the
quoted wording roughly occurs in the cached text. It does not determine whether
the speaker is right or the surrounding claim follows.

\paragraph{Tacit expertise and knowledge elicitation.}
Research on expert knowledge begins with the observation that people can know
more than they can fully state~\citep{polanyi2009tacit}. Cognitive task analysis
uses structured retrospective interviews and records of specific events to
draw out the cues and decisions behind proficient
work~\citep{hoffman1998critical, crandall2006working}. These methods go beyond
summarizing documents. \mimeo{} does not use them or claim to capture tacit
knowledge; it compiles a public record. A future system could combine both
inputs. Recent work on scientific taste follows a different approach. It learns
a community-level preference for potentially influential research ideas from
citation signals~\citep{tong2026taste}. An expert skill would instead try to
represent one person's selection criteria and make them inspectable in context.

\paragraph{LLM judges and persona evaluation.}
Researchers increasingly use LLM-as-judge as an evaluation
measure~\citep{zheng2023judging}. These judges are known to be sensitive to
answer position~\citep{shi2024judging} and to favor their own
output~\citep{wataoka2024selfpreference, panickssery2024selfpref}. JudgeBench
also shows that models that express preferences well can fail on comparisons
with a correct answer~\citep{tan2025judgebench}. PersonaGym uses LLM evaluators
to score how well a model stays in character~\citep{samuel2024personagym}.
InCharacter measures how well a persona is followed through psychological
interviews~\citep{wang2024incharacter}. CoMPosT finds that LLM simulations of
demographic personas drift toward caricature~\citep{cheng2023compost}.

Our results show that this measure is unstable. Four judge models
scored identical prompts and answers. Two rank \mimeo{} as \emph{less} like the
expert than the one-line persona. Their pooled shares fall significantly below
parity, but no individual comparison survives Holm correction within its
six-test family. The other two judges find no pooled effect on the same text.
This pattern is consistent with stereotype-based judgment, but without an
outside key it does not establish caricature as the cause. On the suite whose
tasks carry their own materials, judgment-by-judgment agreement across the
panel is $\alpha{=}0.18$. When judge labels are shuffled within each pair, the
differences between judges are larger than expected if judge identity did not
matter. No judge places the \mimeo{} profile significantly \emph{above} the .5
no-preference mark. This is a failure to replicate, not a reversal.
Appendix~\ref{app:discussion} reports this instability alongside the measures
with an outside key. It does not treat any one judge's score of how much an
answer sounds like the expert as ground truth.

\paragraph{Person-grounded generation and digital twins.}
Generative Agents established believable simulations of human
behavior~\citep{park2023generative}. Character-LLM~\citep{shao2023characterllm}
and RoleLLM~\citep{wang2023rolellm} build role-playing agents from character
profiles and evaluate whether the dialogue holds up. Surveys now cover
role-playing language agents~\citep{chen2024rpla}, including whether these
agents make decisions their persona would make~\citep{xu2024destiny}.
Digital-twin research simulates the behavior of one specific person or
character. BehaviorChain, for example, extracts fictional and nonfictional
characters from fiction and biographical literature because real-world
behavioral data are scarce~\citep{behaviorchain2025}. Those systems aim to
produce believable conversation. \mimeo{} instead produces a portable
description of how an expert reasons. A working agent can use that description
as a file rather than a custom model.

The closest system is COLLEAGUE.SKILL~\citep{colleagueskill2026}, which distills
a target person's materials and interaction traces into versioned skill
packages. The two systems overlap more than they differ. Both emit
\texttt{SKILL.md} bundles under the Agent Skills standard, are open source, and
ship a public gallery of generated profiles. They differ first in their input.
COLLEAGUE.SKILL is built around a colleague the user works with. Its main path
takes user-supplied material, with collectors for chat, email, and document
exports. It also includes a public-figure preset that runs a research pass over
first-person writing and long-form interviews. Public figures are therefore in
scope for both systems, but \mimeo{} begins with a bare name and performs the
discovery itself.

The second difference is the quotation stage. \mimeo{} checks clustered
quotations of at least 20 normalized characters against the cached source text
and drops those that do not match. COLLEAGUE.SKILL's published description
reports coverage and grounding-link checks rather than matching each quotation
against the source. Appendix~\ref{sec:extrinsic-ablations} shows why quotation
matching matters when a system puts words in a named person's mouth. \mimeo{}
also emits \texttt{AGENTS.md} alongside \texttt{SKILL.md}, and our study
evaluates it in one coding-agent harness. These are differences in scope rather
than in kind.

\section{Pipeline Details}
\label{sec:system}

\begin{figure}[t]
  \centering
  \includegraphics[width=\textwidth]{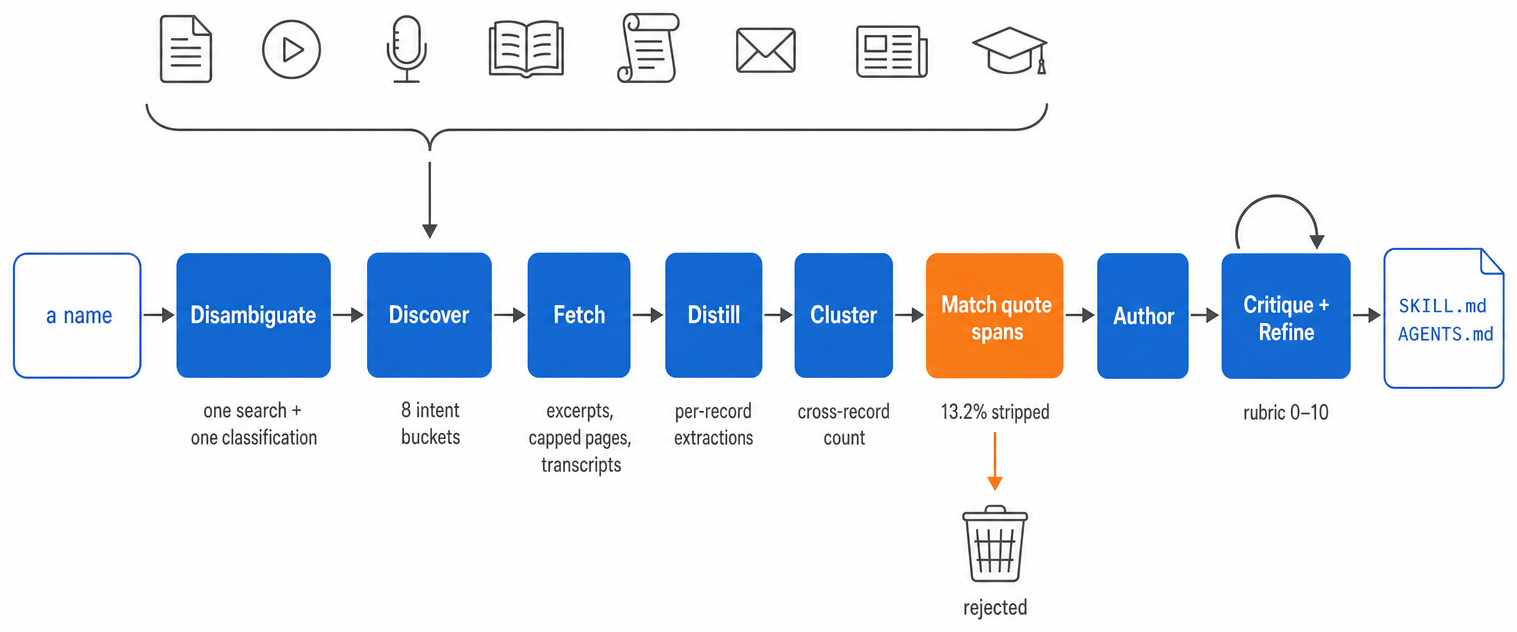}
  \caption{The \mimeo{} pipeline. It first checks whether the name is ambiguous
  and requests disambiguation if needed. It then searches eight kinds of
  material and caches search excerpts, extracted page text, or transcripts.
  The pipeline turns this material into labeled items, groups recurring ideas,
  and ranks them by the number of source records supporting each one. A text
  matcher drops quoted wording it cannot find in the cached source (13.2\% in
  our logged runs). An editor model critiques and revises the resulting artifact.
  ``Source'' here means a retrieved record, not necessarily an independent
  publication.}
  \label{fig:pipeline}
\end{figure}

\mimeo{} is a Python command-line tool. Running
\texttt{mimeo "Richard Feynman"} produces a ready-to-install skill directory.
The pipeline (Figure~\ref{fig:pipeline}) has eight stages and an optional
portrait stage, which is off by default (Appendix~\ref{sec:ethics}). Each stage
caches its output under a fingerprint of its inputs. If a run is interrupted
or reconfigured, it repeats only the stages whose inputs changed. The prompts
treat all fetched text as untrusted data. A URL-safety layer rejects addresses
that carry credentials, point into private networks, or return oversized
responses.

\subsection{Identity disambiguation}
A name can refer to several notable people. Silently combining an economist and a
basketball coach would produce the wrong artifact. Before discovery, \mimeo{}
uses one web search and one LLM classification call to decide whether the name
is ambiguous. In an interactive run, the user picks a candidate. A
non-interactive run stops and prints the exact \texttt{--disambiguator} flag to
pass. The choice is cached separately for each model.

\subsection{Source discovery}
Discovery runs eight searches in parallel, one per kind of material: essays,
talks and lectures, interviews, podcasts, frameworks and principles, books,
papers, and letters or archives. The searches use 27 query templates in total.
These categories cover modern figures, whose work appears in podcasts and blog
posts, as well as historical figures, whose record is in journals and letters.
The pipeline deduplicates results by normalized URL. If more than
\texttt{max\_sources} (default 25) remain, an LLM scores each source in $[0,1]$
for how central it is to the person's public record and keeps the top sources.

\subsection{Fetching}
The pipeline tries three methods for obtaining web material, in order: excerpts
from the search API, a dedicated extraction API, and a local reader. Once the
combined search excerpts exceed 2{,}000 characters, the pipeline accepts them
without fetching the page body. Extracted web text is capped at 50{,}000
characters, so the cache contains evidence from a source but not always the
whole source. YouTube sources use caption transcripts. An optional mode
transcribes podcast audio locally with Whisper. A separate optional
deep-research stage can add a research-agent report as an extra source.

\subsection{Per-source distillation}
The authoring model turns every cached record into labeled items: principles
(statement, rationale, supporting quote), frameworks (name, when to apply,
steps), mental models, heuristics, signature quotes, and anti-patterns. Each
item carries the id of its source record. Records longer than 50{,}000
characters are split at paragraph boundaries (target 40{,}000, overlap
2{,}000), distilled in parallel, and merged by normalized keys. In the pinned
implementation, this path applies only to long transcript or audio records
because the web fetcher has already capped page text at 50{,}000 characters.
Book-length web sources can therefore be cut short without warning. Because the
experiments ran against the pinned release, we keep this behavior and describe
what the code does rather than what it was meant to do.

\subsection{Cross-source clustering}
The pipeline combines items from different records and merges duplicate
concepts. Each merged item keeps a list of its source records. The number of
records supporting an item determines its rank. Normalizing URLs removes exact
duplicates, but it does not catch syndicated copies, mirrors, or two records
made from the same talk. The count is useful for ranking, but it does not prove
that several independent sources support an item. Repeated ideas usually rank
above one-off remarks despite this limit. The pipeline clusters large corpora
in batches that fit the prompt-size budget, then merges the batch results in
memory using a fixed rule (descending frequency, alphabetical tiebreak).

\subsection{Quotation-span matching}
The pipeline checks every clustered quotation against the cached text of its
listed records. It normalizes typography (smart quotes, dashes, whitespace, and
case), first tries an exact substring match, and then falls back to a fuzzy
window match at a 0.82 similarity threshold. It removes anything below that
threshold and writes it to an audit file. The pinned implementation accepts
spans shorter than 20 normalized characters automatically, without matching
them, because short spans are hard to match reliably. None of the 311 spans in
our logged sample was that short. A match shows only that the wording is
present in the cached record. It does not show that the named person said it,
that the surrounding claim follows, or that the record belongs to the right
person. We call this outcome \emph{attested} rather than \emph{verified}. Work
on source-grounded generation makes the same distinction between the presence
of a citation and whether that citation supports the
claim~\citep{liu2023verifiability, gao2023alce}.

\subsection{Authoring}
The authoring stage converts the filtered corpus into one or both target
formats: a
\texttt{SKILL.md} with YAML front matter and seven \texttt{references/} files
(principles, frameworks, mental models, heuristics, anti-patterns, quotes,
sources) following the Agent Skills anatomy~\citep{anthropic2025skills}, and/or
a single always-on \texttt{AGENTS.md}~\citep{agentsmd2025}. The authoring prompt
requires a source id for every claim, but the implementation does not enforce
that requirement. Appendix~\ref{sec:intrinsic} finds four identifiers in the
gallery that point at nothing. Our ablation also finds one matched quotation
credited to the wrong listed source.

\subsection{Critique and refinement}
A critic pass scores the authored artifact from 0 to 10 against a checklist and
lists the problems it finds by category (voice, duplication, unattributed
claims, vagueness, structure, coverage). \mimeo{} uses the critique as
editorial notes, rewrites the artifact, and scores it again. This loop stops
when the score reaches a threshold (default 8) or the revision budget
(default 2) runs out. It always keeps the best-scoring draft and records the
full sequence of scores.

\subsection{Caching, reproducibility, and telemetry}
Every stage's cache key is a SHA-256 fingerprint of the stage inputs, the
prompt-file contents, and the output schema. Editing a prompt therefore
invalidates only the stages that depend on it. A machine-readable run summary
records the number of sources at each stage, the matching statistics, the score
history, and the time each stage took. The telemetry patch we applied to the
pipeline for this study also records token counts, API call counts, and cost.

\section{What the Pipeline Produces}
\label{sec:intrinsic}

We study the pipeline in two ways. First, we examine the 22-expert gallery
committed with the pinned toolkit release.\footnote{\url{https://github.com/K-Dense-AI/mimeo}}
Most gallery runs used \texttt{--format both}, and the gallery includes optional
generated avatars. It was therefore not produced entirely with default settings. Second, we
re-run eight experts end to end and log their tokens and API calls. The gallery
was written by the pinned toolkit's default model,
\texttt{google/gemini-3.1-pro-preview}; the re-runs use the model we pin for the
rest of the paper. The two subsections therefore cover the same pipeline but
not the same authoring model. The counts below describe the pipeline's
structure. The per-run telemetry and quality numbers in
Appendix~\ref{sec:intrinsic-reruns} use the model we pin for the rest of the
paper.

\begin{figure}[t]
  \centering
  \includegraphics[width=\textwidth]{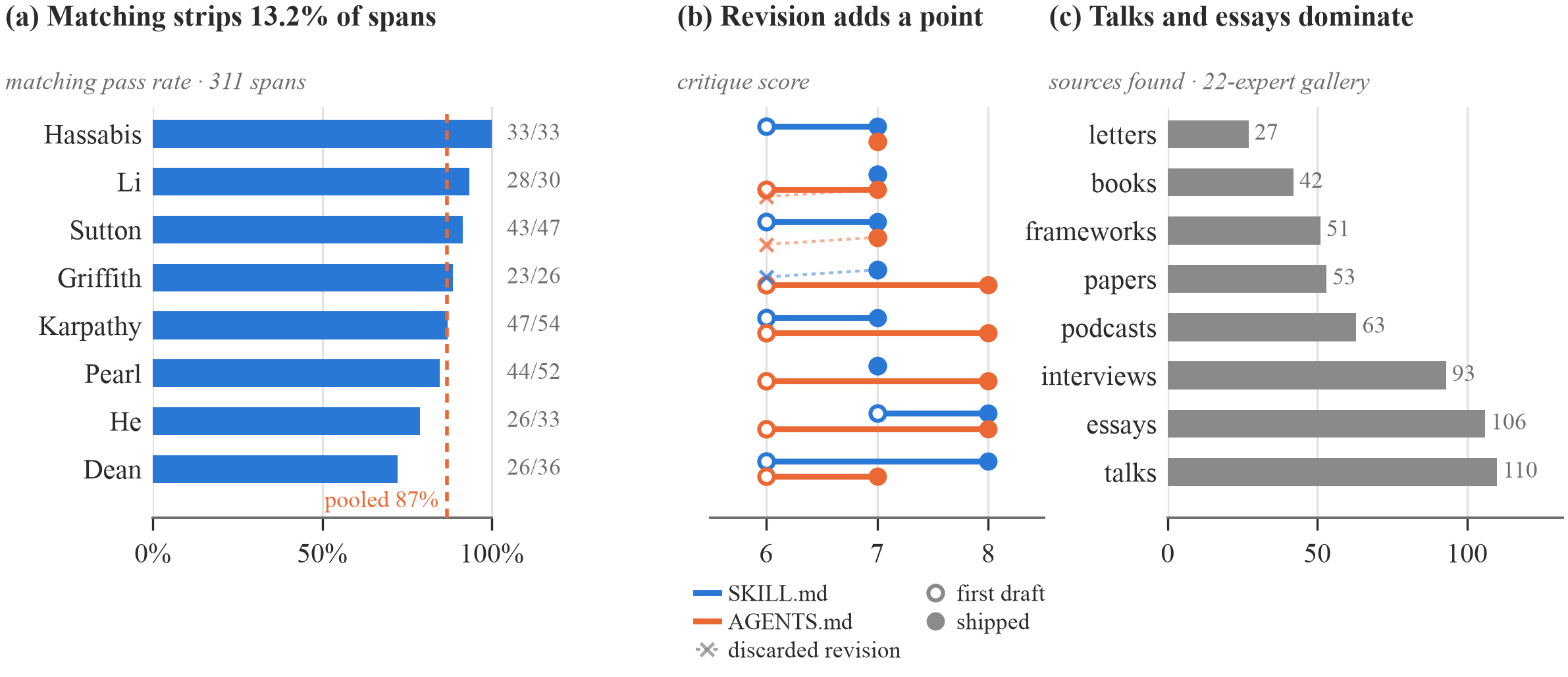}
  \caption{\textbf{Both editing stages change their outputs.}
  (a)~Quotation matching removed 41 of the 311 spans extracted across the eight
  logged re-runs (13.2\%; pooled pass rate marked), and rejected at least one
  quote for seven of the eight experts;
  Appendix~\ref{sec:extrinsic-ablations} shows what ships when this stage is
  turned off. (b)~Critique scores for the first draft (hollow) and the shipped
  file (filled) of each expert's two artifacts. The loop gains $+0.75$ points on
  average for \texttt{SKILL.md} and $+1.25$ for \texttt{AGENTS.md}. Crosses mark
  the three files whose \emph{last} revision scored below an earlier draft.
  Keeping the best draft shipped the earlier one instead, so no run ends below
  where it began. (c)~The mix of search categories across the 545 sources of the
  22-expert gallery. Panel (b) uses the expert order from panel (a).}
  \label{fig:intrinsic}
\end{figure}

\subsection{The generated artifacts}
An average gallery bundle (\texttt{SKILL.md}, \texttt{AGENTS.md}, and seven
reference files) contains 6{,}453 words, with a range of 4{,}007 to 8{,}738.
This total consists of a $\sim$970-word \texttt{SKILL.md}, a
$\sim$2{,}070-word \texttt{AGENTS.md} (across the 20 of 22 experts generated
with \texttt{--format both}), and $\sim$3{,}600 words of references. The median
bundle has 9 principles, 5 frameworks, 7 mental models, 8 heuristics, 8
anti-patterns, and 14 signature quotes. The pipeline uses nearly all of its
default 25-source budget for each expert (mean 24.8, minimum 24). Most of the
discovered sources are talks, essays, and interviews
(Figure~\ref{fig:intrinsic}c). Letters contribute little except for people
whose work has historical archives. The LLM scores for the 544 ranked sources average
0.85; one of the 545 discovered sources was not scored.

\subsection{Attribution structure}
The files carry 113 source references per expert on average. The median expert
cites all 25 discovered records at least once (mean proportion 0.96, minimum
0.52). The pipeline links two or more records to 37\% of the distilled items
(mean 1.55 records per item, maximum 7). These counts show which sources are
linked to each item, not whether each source supports it. We did not ask expert
human raters to check whether each cited record supports the claim beside it.
The bookkeeping also has errors. Four of the 22 gallery bundles each contain
one identifier that points at nothing, and
Appendix~\ref{sec:extrinsic-ablations} finds a matched quote credited to the
wrong source. Calling these bundles ``verified expert skills'' would claim more
than the quotation matcher can show.

\subsection{Logged re-runs}
\label{sec:intrinsic-reruns}
We re-ran eight experts from machine learning, causal inference, and biological
engineering.\footnote{Andrej Karpathy, Richard S.\ Sutton, Kaiming He, Jeff
Dean, Judea Pearl, Linda Griffith, Fei-Fei Li, Demis Hassabis; pipeline
defaults except \texttt{--format both} and the authoring model we pin
throughout, \texttt{google/gemini-3.6-flash}.} A telemetry patch records token
use and API call counts for each run. A complete run averages 37.8 LLM calls,
258k prompt tokens, and 107k completion tokens. It also averages 8 search
calls, 8--16 extraction calls, and 7.1~minutes of elapsed time. The API-call
and token totals do not measure local compute or include the optional
transcription and portrait stages. The elapsed time covers only the logged
configuration. These figures therefore do not capture the full cost of
operating the pipeline.

Two stages change their outputs. Quotation matching
(Figure~\ref{fig:intrinsic}a) rejected 13.2\% of extracted quotation spans
overall, with per-expert pass rates from 0.72 to 1.00;
Appendix~\ref{sec:extrinsic-ablations} shows what ships when this stage is
turned off. The critique-and-revise loop (Figure~\ref{fig:intrinsic}b) raised
the critique score by $+0.75$ points on average for \texttt{SKILL.md} and
$+1.25$ for \texttt{AGENTS.md}. Revising did not always help: in 3 of 16 files,
the last revision scored below an earlier draft. Because the loop always keeps
the best-scoring draft, it shipped the better earlier one.

Two limits matter when reading this gain beside the near-ceiling E2 result.
First, the critique score is not an independent measure of quality. The
authoring model scores its own draft against the checklist that the loop
optimizes. Because the loop keeps the best draft, the score cannot finish below
where it started. Three of the eight \texttt{SKILL.md} files show no score gain.
We report how the loop performs on its own checklist, which does not reflect an
independent quality assessment. Appendix~\ref{sec:extrinsic-ablations} also shows that the
checklist cannot see how the corpus is organized, which further limits what the
score can tell us. Second, most files exhaust the revision budget below the
pipeline's default quality bar of 8.
Shipped scores were 7,\,7,\,8,\,8,\,7,\,7,\,7,\,7 for \texttt{SKILL.md} and
8,\,7,\,8,\,7,\,8,\,8,\,7,\,7 for \texttt{AGENTS.md}: 6 of 16 cleared the bar.
The four \texttt{AGENTS.md} files used in every behavioral experiment have
these scores: Karpathy 8, Sutton 7, He 8, Dean 7. By the pipeline's own
checklist, half of these four experimental files are a point short of the
target, which matters when interpreting Appendix~\ref{sec:extrinsic}. They do
contain the material tested later: each answers all five single-record questions about its expert; collectively, they answer all 20. BM25 keyword
search given only the question answers 15--17 of the 20 questions. This shows
that the content is available, not that the writing around it is good.

\section{Do Expert Files Change Agent Behavior?}
\label{sec:extrinsic}

A profile grounded in sources can be useful without changing what an agent
does. We test the stronger claim here. Once the file is loaded, does the agent
notice different problems, give better advice, or leave a recognizable trace
of the named expert in its answer?

\subsection{Design}
\label{sec:extrinsic-design}

\paragraph{Tasks.}
We wrote eight software-engineering tasks that call for judgment
(Appendix~\ref{app:tasks}): code review, architecture critique, debugging
strategy, refactoring planning, experiment design, API design review,
performance investigation, and incident postmortem. None of the tasks names an
expert. Four have fixed lists of 8, 3, 6, and 4 planted issues, 21 in
total. We wrote these
lists before running any condition and kept them fixed across conditions. The
tasks are ordinary engineering scenarios rather than problems tailored to an
expert, and they match settings where people install an always-on context file.
They are not, however, the easiest way to test whether Sutton's or He's
published work transfers.
Appendix~\ref{sec:regimes-application} adds expert-relevant
scenarios.

\paragraph{Conditions.}
Each run starts a fresh, unattended Claude Code session using
\texttt{claude-sonnet-5}. The session receives only one kind of project
context: \textbf{no skill}, no file; \textbf{one-line persona}, an instruction of about 33 words
that names the expert; \textbf{from-memory profile}, an \texttt{AGENTS.md} of
similar length and format to the \mimeo{} file, written by the same Gemini
authoring model without sources in front of it; or \textbf{\mimeo}, the
\texttt{AGENTS.md} built from the corpus. We test each of the three persona
conditions with four experts (Andrej Karpathy, Richard Sutton, Kaiming He, and
Jeff Dean). We generate the no-skill answer once per task. The initial grid has
$8\times(3\times4+1)=104$ runs.

One answer per combination cannot show how much the answers vary. We therefore
generated two more answers for every combination on the four keyed tasks. With
the original pass, this gives three separate answers per task, condition, and
expert: 156 keyed answers, 12 with no skill and 48 per persona condition.
Because we built each artifact only once, the conclusions apply to these four
profiles as built.

\paragraph{Measures.}
Each grader receives a numbered issue list and marks every issue
\emph{found}, \emph{partial}, or \emph{missed}. Coverage counts a partial
finding as half credit, and we also report found-only rates. Two graders from
different developers score every answer. For task-level contrasts, we first
average across experts and generations. This leaves four independent units,
one for each task.

Judges also compare pairs of answers for advice quality and for how well they
match the named expert. Each pair is judged twice, once in each order. We treat
these measures as secondary because the design can detect only large
preferences, and raters often disagree with themselves after the answer order
is swapped. Appendix~\ref{app:discussion} reports the full four-judge result.

\begin{figure}[t]
  \centering
  \includegraphics[width=\textwidth]{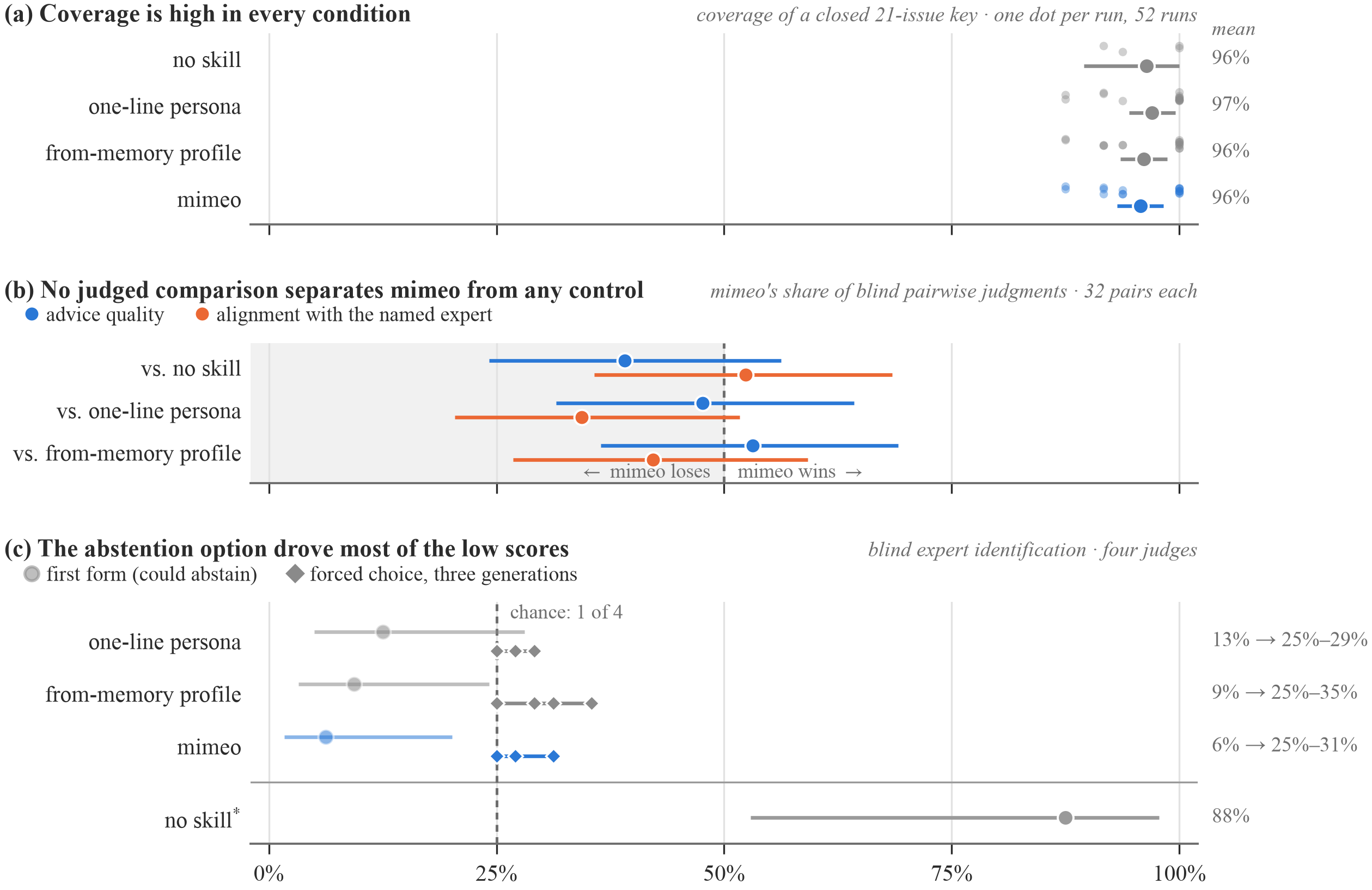}
  \caption{\textbf{Engineering coverage is non-inferior at the ceiling; advice
  quality remains unresolved.} (a)~Coverage of the fixed issue list in the first
  pass, primary grader; every condition is at 96--97\%, which rules out a large
  loss but leaves almost no room to observe improvement. Two further generations
  for each keyed combination reproduce it, and the second grader runs 1--2 points
  lower in every condition (Table~\ref{tab:e2-repeats}).
  (b)~Pairwise judgments from the primary judge; no quality comparison survives
  correction, and the four-judge panel does not support any single judge's
  ordering on expert alignment (Section~\ref{sec:discussion}). (c)~The
  identification question in its first form, which allowed abstention (faded
  circles, 32 answers each), beside the shuffled forced choice used for every
  reported result (one diamond per judge, 48 repeated-task answers each). The
  very low scores on the first form measured the judge's willingness to answer
  ``none,'' not the persona. $^{*}$The no-skill row exists only for the first
  form, where ``none of them'' is the correct answer for its 8 answers;
  a forced choice among four expert names has no counterpart for it. It shows
  the instrument can detect a signal when one is present.}
  \label{fig:extrinsic}
\end{figure}

We also ask each judge which expert was loaded. The first version included a
``none of them'' option and always listed the names in the same order.
Abstention and order drove much of the score. The reported measure instead
forces a choice among four names, shuffled for every answer by a fixed rule.
Four judges score the same answers, and chance is 25\%. We report both
versions in Figure~\ref{fig:extrinsic}c because the question changed, not the
answers, and this accounts for most of the difference between them.

\subsection{Results: non-inferior coverage at the ceiling; quality unresolved}
\label{sec:extrinsic-results}

\begin{table}[t]
  \centering\small
  \setlength{\tabcolsep}{5pt}
  \begin{tabular}{lcccc}
    \toprule
    & no skill & one-line persona & from-memory profile & \mimeo{} \\
    \midrule
    Coverage, primary grader & .969 & .960 & .966 & .960 \\
    Coverage, second grader  & .950 & .942 & .956 & .949 \\
    Found only, primary      & .938 & .936 & .943 & .933 \\
    Forced identification, four-judge range & --- & .250--.292 & .250--.354 & .250--.313 \\
    \bottomrule
  \end{tabular}
  \caption{Three separate generations for each keyed combination. Coverage uses 12
  no-skill answers and 48 answers per persona condition. Forced
  identification has $n=48$ answers per condition for each judge, and its ranges
  span the four fixed judges. Intervals formed by resampling task means include
  the 25\% chance rate for every judge in the \mimeo{} and one-line persona
  conditions. One of four judge results for the from-memory profile excludes
  it; none of these
  results establishes equivalence to chance.}
  \label{tab:e2-repeats}
\end{table}

\paragraph{Coverage holds across generations, but the task is already at its ceiling.}
Under the primary grader, the three-pass means are .969 with no skill, .960
with the one-line persona, .966 with the from-memory profile, and .960 with
\mimeo{}. The second grader gives .950, .942, .956, and .949. The two graders
agree on 95.1\% of 819 issue verdicts, and on 96.2\% of the simpler
found-versus-not decision. Repeating a condition moves its mean by at most 3.5
points.

For each comparison, we average generations and experts within task, then test
the four task differences against a $\pm.10$ margin. This equivalence test asks
whether the difference is small rather than whether it is nonzero. All six
tests (two graders by three baselines) meet that standard ($p\le .0053$), and
every 95\% interval lies within $[-.049,+.059]$. We fixed the margin after
seeing the original single-generation coverage numbers but before analyzing
the two further generations. It therefore constrains the repeated analysis
without being tuned to those results. It is the smallest change that would
alter what we would tell someone to deploy. With only four tasks, this tells us
how sensitive the result is to a difference of this size; it does not show how
the result applies to tasks in general.

TOST tests both directions and rules out differences larger than ten points in
either direction. The practical meaning differs by direction because an agent with
no skill already covers about 95--97\% of the list. Even a perfect answer could
gain only 3--5 points. The test rules out a ten-point loss in coverage, but it
cannot show that profiles never help on harder tasks. We therefore call this
ceiling-limited non-inferiority, not proof that behavior is unchanged.

\paragraph{With names shuffled, judges find little trace of the \mimeo{} expert.}
On the repeated keyed tasks, judges name the \mimeo{} expert correctly .250 to
.313 of the time, the one-line persona .250 to .292, and the from-memory profile
.250 to .354. Intervals formed by resampling task means include the 25\%
reference for every judge in the \mimeo{} and one-line persona conditions. One
of the four judge results for the from-memory profile excludes it, at
[.29,.42], but that isolated signal does not replicate across judges. Some
judges find a weak from-memory signal in the original eight-task pass, but the
repeated keyed answers do not reproduce it. This changes how the original
6--13\% scores should be read. Those very low numbers mostly measured a judge's
willingness to answer ``none,'' not negative evidence about persona content.
The narrower result is that these four \mimeo{} profiles leave a weak,
unresolved identification signal on repeated tasks that carry their own
materials.

\paragraph{Pairwise quality stays unresolved.}
Across the original 32 pairs per comparison, \mimeo{} takes .39--.53 of the
vote against the three controls. Each pair is judged in both orders, and no
comparison is significant after Holm correction for multiple testing. At the
observed tie rate, a conditional power calculation for the exact sign test
reaches 80\% only when the vote share is around .68--.72. In other words, this
sign test is likely to detect a preference only when that preference is large.
The calculation does not cover the whole design and cannot rule out a modest
preference. Comparisons against the shared no-skill answer say even less:
collapsing to eight tasks gives $p=.45$ for quality. We do not use pairwise
quality to support the coverage result.

\subsection{The on-demand deployment}
\label{sec:extrinsic-ondemand}

In the always-on experiment, the profile is available throughout every run,
giving it every chance to affect the answer. Agent Skills work differently.
The agent first sees a short description, then decides whether to open
\texttt{SKILL.md}. We installed the same four bundles under
\texttt{.claude/skills/}, reran the four keyed tasks for 16 runs in total, and
recorded the event stream.

The agent loaded the skill in 5 of 16 runs (31\%, 95\% CI [14\%, 56\%]): three
Karpathy runs, one Dean run, one He run, and no Sutton run. This small arm
cannot establish differences between experts, but it shows that deployment
adds another bottleneck. A faithful description of Sutton's research agenda
does not necessarily look relevant to a code review.

Coverage is .979 for the on-demand deployment and .957 for the original
always-on pass. After averaging the four expert cells within each of the four
tasks, the paired difference falls inside the same ten-point margin
($p=.0017$). Eleven of the sixteen runs never loaded the file, so this is a
result about the deployment rather than the contents. Coverage among the five
runs that did load it is .975, but five observations are too few for a
separate claim.

\subsection{Pipeline ablations}
\label{sec:extrinsic-ablations}

\paragraph{The quotation filter catches real failures.}
We regenerated three experts with matching disabled and checked the quotations
in the shipped files against the cached source text. With filtering on, 0 of 34
spans were unmatched; with it off, 2 of 36 were invented sentences presented as
direct quotations of living people. This small comparison shows that the
failure can happen, not how often. One quotation that matched was still
assigned the wrong source id. Speaker attribution needs a human audit before
these files can be called verified.

\paragraph{The critique checklist cannot see clustering.}
Authoring from ungrouped extractions used 140, 153, and 54 items, against 58,
62, and 29 after clustering, a $1.9\times$--$2.5\times$ consolidation. The
resulting files received the same self-critique score (7 versus 7). The
checklist measures the surface of the writing, not the structure of the corpus
behind it. The revision loop improves its own score, but we did not measure the
behavioral value of clustering.

\section{Where the Corpus Matters}
\label{sec:regimes}

The engineering study leaves two questions open. Can the file supply details
the model does not already know from training? And when does the persona remain
visible in an answer? We examine direct recall, retrieval, application, and
open-ended steering separately rather than treating them as one effect.

\subsection{E4: direct questions about the corpus}
\label{sec:regimes-knowledge}

\paragraph{Design.}
We wrote 40 questions about statements the four experts have made on the
record, ten per expert. Five per expert concern \emph{canonical} material
repeated in many places. The other five concern \emph{tail} material found in
a single record; the body calls these the single-record questions. Here, tail
describes where the material sits in the corpus,
not how broad the question is: 19 of its 20 probes ask for a signature
quotation or phrase, so this is mostly a test of exact recall. Every answer key
comes from the cached corpus used to build the \mimeo{} file.

One consequence of that construction deserves to be stated plainly. We selected
each probe by picking a distilled item out of the same clustered corpus the
\mimeo{} file was written from, and the key is that item. The material an answer
has to produce is therefore material the \mimeo{} file was built to carry, and
Appendix~\ref{sec:intrinsic} confirms the four files do carry it. A ceiling
score for \mimeo{} on this instrument is close to guaranteed by the design and
is not evidence about the pipeline's editorial choices. What the instrument can
inform is the comparison among conditions that were not built from the key: the
three closed-book controls, which show how much of this material the model holds
without the corpus, and BM25 retrieval, which reads the same cached records and
is the only condition with the same access to the answer.

We collected one response to every question under each of the four original
conditions. Four graders, blind to condition, scored each answer against the
fixed key as correct, partial, incorrect, or abstained. A separate text matcher
checked quoted spans against the cached source.

This design gives \mimeo{} an open book and every original control a closed
one. We therefore added a fifth condition: BM25 keyword search over the same
cached records~\citep{robertson2009bm25}, given only the question. The
retriever ranks 2{,}400-character chunks and places up to five in context,
capped at the length of the matching \mimeo{} file. It retrieves a key source
for 36 of the 40 questions. This is a deliberately simple retrieval baseline,
not a tuned RAG system.

\paragraph{Answering from memory fails on obscure wording.}
Under the primary grader, \mimeo{} answers all 40 questions, including 20 of 20
in the tail. The no-skill, one-line persona, and from-memory profile conditions
answer 40\%, 25\%, and 45\% of the tail. All four graders score \mimeo{} at
20/20; no closed-book control exceeds 10/20. Agreement among graders on correct
versus not is strong (Krippendorff's $\alpha=.83$).

The closed-book conditions also fail in different ways. Declining to answer is
honest; naming a position the record does not document is not. On the 20 tail
questions, the no-skill agent is never graded incorrect under any of the four
graders, and neither is \mimeo{}. The two persona conditions are graded incorrect: the one-line
persona on 3, 1, 2, and 2 of 20 answers across the four graders, and the
from-memory profile on 3, 3, 4, and 3. BM25 retrieval is graded incorrect once under one grader
and never under the other three. On questions where it has nothing to go on, a
persona file written without sources therefore moves the agent from declining to
naming the wrong position. This pattern is specific to the tail.
On the widely repeated canonical questions, the no-skill agent errs too (5\%
under the primary grader) because there it does have something to say. This is
the sharpest behavioral difference we observe between a grounded and an
ungrounded profile. It is also the only measure on which \mimeo{} clearly beats
the cheap alternative rather than matching it.

\begin{figure}[t]
  \centering
  \includegraphics[width=\textwidth]{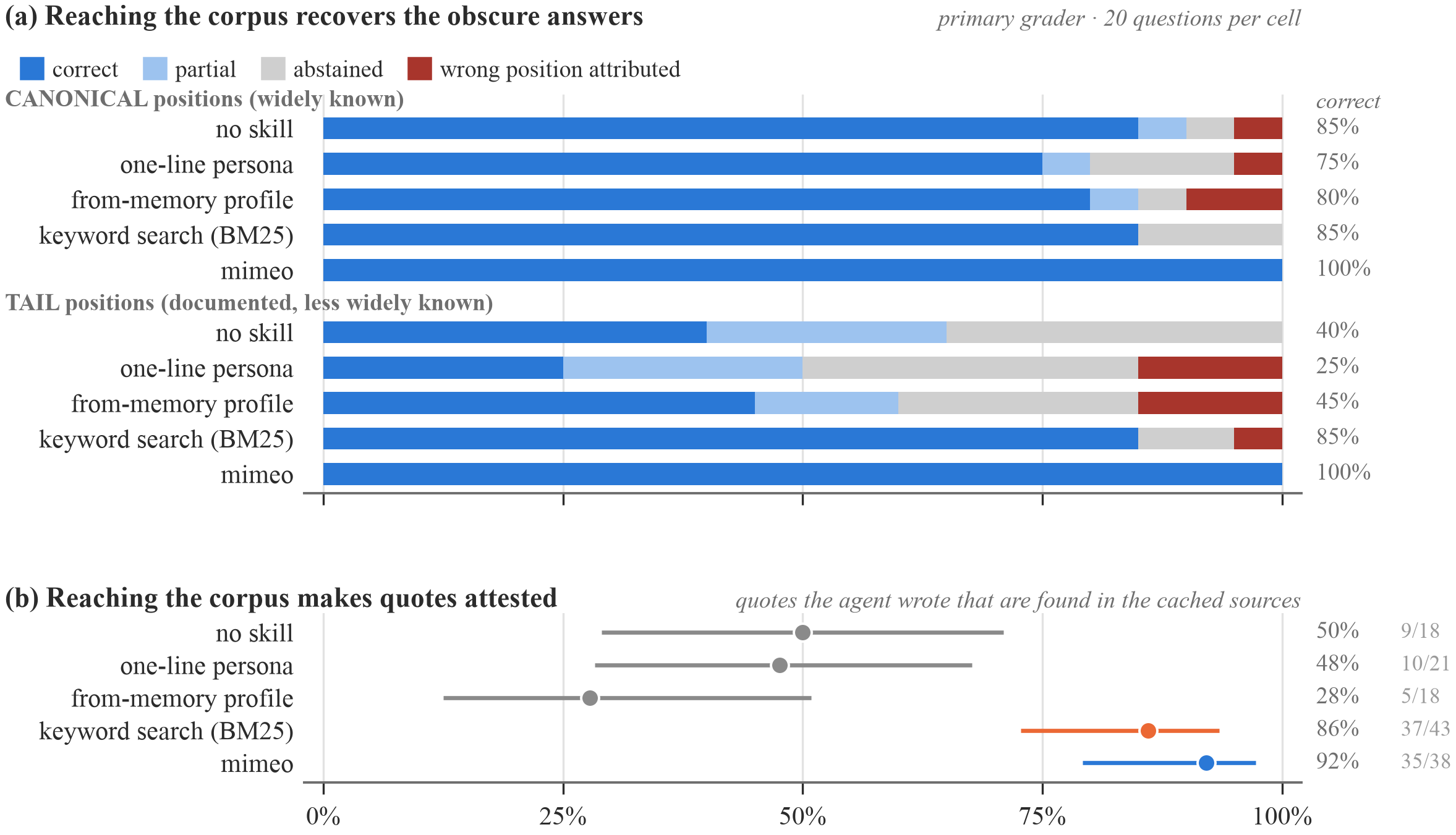}
  \caption{\textbf{The corpus supplies obscure details, but we cannot yet
  tell whether distillation improves on plain retrieval.} (a)~\mimeo{} answers
  all 20 quotation-heavy questions drawn from a single record; the closed-book
  conditions answer 5--9. Given only the question, a BM25 keyword search over
  the same cached corpus answers 17 under this grader and 15--17 across the
  panel. Its gap from \mimeo{} is not statistically significant under any
  grader. The widely repeated questions are near the ceiling. (b)~The wording
  appears in the cached corpus for 92\% of quoted spans under \mimeo{} and 86\%
  under retrieval, against 28--50\% for the closed-book conditions. Matches
  make quotations attested, not verified.}
  \label{fig:knowledge}
\end{figure}

\paragraph{What the incorrect grade counts.}
The rubric marks an answer incorrect when it attributes a position that differs
from or contradicts the key, and the grader sees a positive key rather than a
list of positions to rule out. Two kinds of failure therefore land in the same
category. Some answers invert the documented distinction. Asked for Karpathy's
line about which cognitive work can be delegated, the one-line persona and the
from-memory profile both put thinking on the non-delegable side, which reverses
it. Other answers name a different plausible position. Asked what Kaiming He
calls the most important discovery of deep learning, both personas answer
residual connections rather than the transferability of deep representations
that the key documents. The second kind is a wrong answer about what the person
said in that record. It is not proof that the person never held the view, and it
is not a measure of invented content. The quotation matching in
Figure~\ref{fig:knowledge}b measures that separately. Nor does the contrast rest
on a probe-level test: zero against three of 20 probes is not significant. It
rests on four graders, scoring independently, recording the same ordering in
both persona conditions.

These results establish an open-book advantage, not a benefit unique to
\mimeo{}'s editorial pipeline. Plain BM25 answers 17/20 under the primary
grader and 15--16 under the other three. Question by question, retrieval misses
three to five answers that \mimeo{} gets right, while \mimeo{} misses none that
retrieval gets right. Yet the exact McNemar tests give $p=.063$--$.25$. With
only 20 probes, the 15--25 point difference remains unresolved. A better
retriever might close it, while a larger study might confirm it.

\paragraph{Access to the corpus improves quotation matching.}
Of 38 quotations generated under \mimeo{}, 35 (92\%) are found in the cached
source. For BM25 retrieval, the rate is 37 of 43 (86\%). The original
closed-book conditions range from 28\% to 50\%. This difference is largely
expected because agents often copy text placed in front of them. Access to the
corpus reduces unsupported quotation, but the result does not show that a
static distilled file beats retrieval when each question arrives. The
unmatched spans are not all inventions. Some are real statements from outside
the 25 cached records, some contain ellipses, and some fall just below the
fuzzy threshold.

\subsection{E4b: less famous experts}
\label{sec:regimes-familiarity}

\begin{figure}[t]
  \centering
  \includegraphics[width=\textwidth]{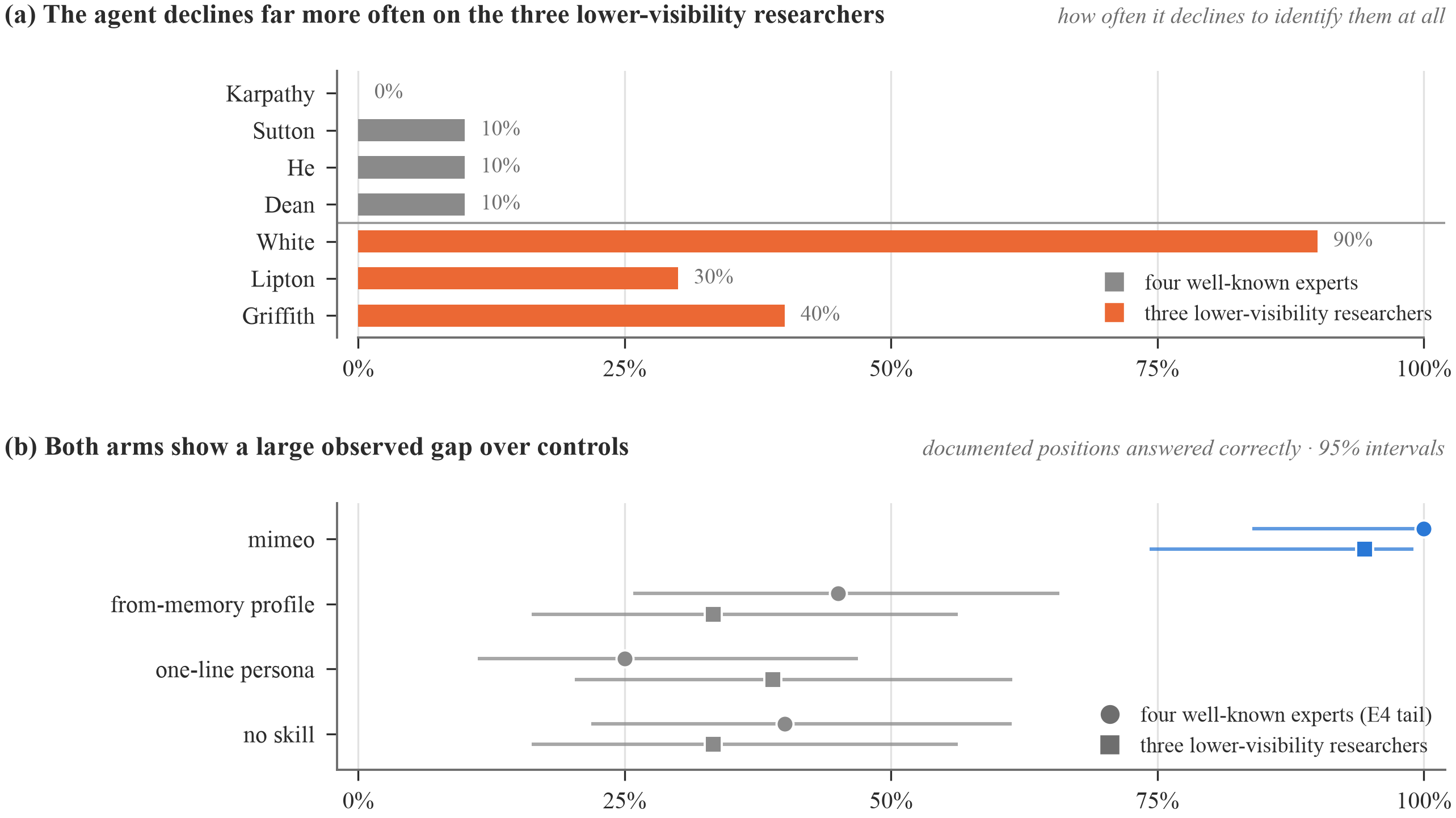}
  \caption{\textbf{The gap over answering from memory holds for less famous
  experts.} (a) The agent declines much more often when asked to identify the
  three lower-visibility researchers. (b) Squares give the share
  answered correctly on the 18 single-record questions for these three
  researchers, with circles repeating E4's four well-known experts for
  comparison; bars are 95\% intervals. \mimeo{} answers 17 of the 18,
  against 6 for the from-memory profile, 7 for the
  one-line persona, and 6
  with no skill. The gap between \mimeo{} and the from-memory profile is
  similar in size to E4's, but the interval around that comparison is wide.
  This arm has no runtime-retrieval condition.}
  \label{fig:familiarity}
\end{figure}

Because the four original experts are unusually prominent, we ran an
exploratory follow-up on Martha White, Zachary Lipton, and Linda Griffith. In a
separate free-recall check, the agent model knows them less well. It declines to
identify the three in 30--90\% of samples, against 0--10\% for the original
four. Each corpus passes a fixed screen for item count, repeated support,
matched quotations, and the share of sources that are directory pages.
Figure~\ref{fig:familiarity} reports the arm.

The \mimeo{} condition answers 17 of 18 questions under both graders. The
from-memory profile, one-line persona, and no-skill conditions answer 6, 7, and
6 under the primary grader. Exact paired tests on the 18 probes give
$p\le .002$, but do not show that the result applies beyond these three
experts. All 13 quotations produced under \mimeo{} match the cached text,
against half under each control. One from-memory answer produces an unattested
quotation about endometriosis that scores .44 against the .82 matching
threshold.

We expected the gap over the from-memory profile to widen as the model became
less familiar with an expert. This arm does not establish whether it does. The
estimated change is +.06 with a 95\% interval of $[-.20,+.33]$ under the
primary grader, and +.01, $[-.25,+.32]$, under the second. These percentile
intervals use 20{,}000 bootstrap draws, resampling the three less-famous and
four original experts separately. With so few expert clusters, the intervals
show how sensitive our result is; they are not estimates for a wider
population.

We report E4b as exploratory. Our preregistration covers only the four-judge
replication; the E4b hypotheses appear in code comments, with no record fixed
before the outcome was known. Appendix~\ref{app:e4b} gives the registration
status in full.

\subsection{E6: applying principles to new scenarios}
\label{sec:regimes-application}

Questions about quotations are a weak test of judgment transfer. Before running
E6, we fixed a local analysis plan and 16 new scenarios, four per expert. None
names the expert. Each asks for a decision based on a principle from the corpus,
not a quotation. Examples include choosing a residual reformulation over more
tuning of a failing 100-layer plain network and choosing a five-to-tenfold
scaling horizon over an unsupported thousandfold one. The five conditions are
no skill, one-line persona, from-memory profile, \mimeo{}, and BM25 retrieval.
To mark an answer correct, four graders require both the expected choice and
the reasoning specific to that principle.

The test hits its ceiling. The primary and second graders mark all 16 answers
correct in every condition. The other two put each condition at 15--16 of 16,
and no answer is graded contrary to the key. The graders agree unanimously on
75 of the 80 condition-by-probe cells. For the two comparisons fixed in the
analysis plan, three graders find no probe where \mimeo{} and the baseline
differ. The remaining grader, \texttt{grok-4.6}, marks one probe correct
only under \mimeo{} for each baseline (exact McNemar $p=1.0$; difference
interval $[0,.19]$ formed by resampling experts). The registered rule that all four graders must agree on
direction is therefore not met. The scenarios turned the corpus principles
into choices that the base model already treated as ordinary good practice.
E6 gives no evidence of judgment transfer and none against it. A useful next
test needs cases where the expert's documented recommendation departs from the
model's default, ideally graded by people who know the expert's work.

\subsection{E5/E5b: open prompts and a paired context intervention}
\label{sec:regimes-openended}

\paragraph{Design.}
E5 contains six short prompts asking for a research agenda, a startup
direction, a course, a hiring loop, a ranking of open problems, or PhD advice.
Each asks for firm choices and attaches no materials. The full condition grid
has 78 runs. Comparing E5 with the E2 task suite would not isolate the effect
of attached materials because the suites also differ in context length, task
domain, and how well each expert fits the task.

E5b provides the paired intervention. It keeps each E5 request word for word
and attaches 6.9--7.1k characters of task-specific evidence and shared
organizational records. The packet names no expert and deliberately leaves
room for several recommendations. We rerun all 78 cells. For identification,
we require a choice among four shuffled names instead of using the original
question, which allowed abstention.

The judge sees the same prompt the agent saw, so the packet lengthens the
judge's context too, from about 400 characters of request to about 7{,}300. The
paired drop therefore combines two changes: what the answer carries, and what
the judge reads beside it. Showing the judge the packet-free request in both
arms would separate them. We report the drop as a change in how visible the
persona is once task material is present, not as an isolated effect on the
answer.

\paragraph{Every persona is visible on short prompts.}
Across the four judges, identification on E5 ranges from .667 to .833 for the
one-line persona, .708 to .958 for the from-memory profile, and .625 to .917
for \mimeo{}. Every 95\% interval formed by resampling tasks lies above the
25\% chance rate. Basing the profile on sources does not make it easier to spot:
the from-memory profile is at least as identifiable as \mimeo{} for every
judge. The persona is visible, but this is not evidence that the corpus
transferred judgment.

\begin{figure}[t]
  \centering
  \includegraphics[width=\textwidth]{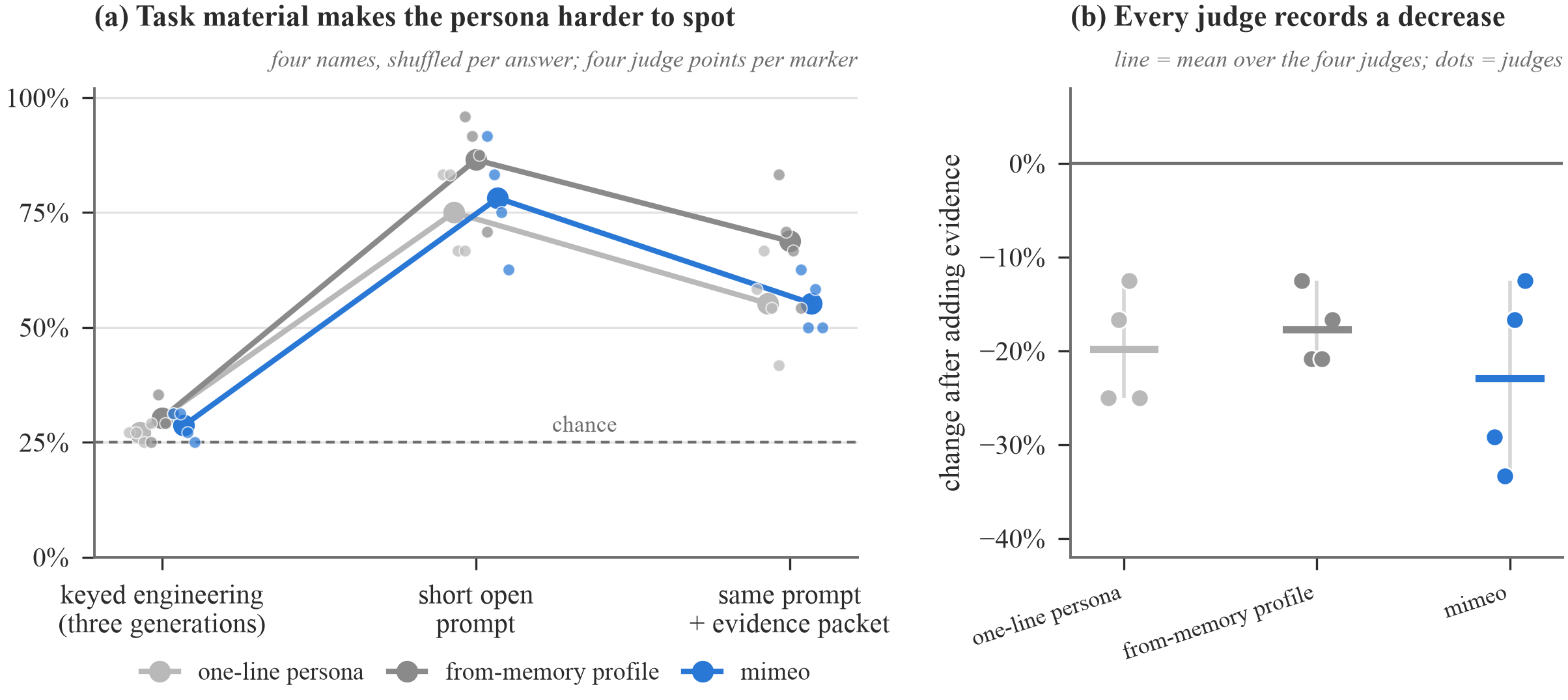}
  \caption{\textbf{Task material weakens the persona without erasing it.} How
  often four judges name the right expert when the four names are shuffled for
  every answer. (a)~The same three personas are near chance on the keyed tasks,
  which carry their own materials. They are easy to spot on short open prompts,
  then fall back partway when the paired evidence packet is attached.
  (b)~Averaged over the four judges, the drops are 20 points for the one-line
  persona, 18 for the from-memory profile, and 23 for \mimeo{}; every judge
  records a drop for every persona.}
  \label{fig:openended}
\end{figure}

\paragraph{Adding materials reduces the signal.}
With the packet attached, the same ranges are .417--.667, .542--.833, and
.500--.625. Every judge records a drop for every persona type. Averaged over
the four judges, the drops are .198, .177, and .229
(Figure~\ref{fig:openended}). Treating the six tasks as
the independent units and averaging judges and persona conditions within each
task gives six negative differences. An exact sign-flip test gives $p=.031$.
The packet does not erase the signal: every E5b point estimate remains above
.25, although the intervals formed by resampling tasks are wide. The data
support the view that task material competes with the persona. They do not show
that adding the packet eliminates steering.

\paragraph{Judged quality gives no panel result.}
The primary judge prefers the no-skill E5 answer over the \mimeo{} answer
(\mimeo{} share .23), as does one replicate judge (.17); the other two give .50
and .61. Because the same no-skill answer is shared across four expert
comparisons, we analyze the six independent tasks instead. This gives $p=.219$
by the sign test and $p=.094$ by Wilcoxon. We therefore drop the claim that
personas reduce advice quality. Under this measure, the experiment shows
neither a reliable benefit nor a reliable penalty.

\subsection{What the studies establish}
\label{sec:regimes-summary}

The results separate four claims that a single ``knowledge, not steering''
verdict would blur. Access to the corpus clearly improves recall of obscure
wording over answering from memory. Simple retrieval when each question
arrives recovers most of that gain, and this sample does not establish an
advantage for the static distilled file. New application scenarios hit the
ceiling in every condition, so judgment transfer remains unmeasured. The
persona is highly visible on short open prompts and about 20 points less
visible once a paired evidence packet is attached. These findings support
\mimeo{} as a compact artifact that carries its sources. They do not show that
installing it transfers an expert's judgment.

\section{Extended Discussion}
\label{app:discussion}

\paragraph{What transfers?}
Section~\ref{sec:discussion} gives the four main conclusions. This appendix
examines the questions behind them: why task materials weaken the persona, what
a static artifact offers beyond retrieval, what the self-critique score can
show, and why measures based only on a judge's sense of expert voice vary.
Throughout, we distinguish knowing things from sounding like someone and
deciding like someone.

\paragraph{Why is the persona weak on engineering tasks?}
The paired E5/E5b intervention points to one explanation: task materials
compete with the file for the agent's attention. Adding 6.9--7.1k characters
to the same six requests lowers identification for every persona under every
judge. Identification remains well above the E2 level, where the tasks
include their own materials, so context length is not the whole story. The fit
between expert and task also matters. A Sutton profile has distinctive things
to say about an AI research agenda and less to say about an API review. The
base model also contains much of the common engineering advice. Across three
separate generations, every condition covers nearly all planted issues, and
every condition gives the keyed answer on nearly every E6 scenario.

The repeated E2 results cover one commercial agent model, four profiles,
four keyed tasks, and three generations for each combination. On these tasks,
they rule out a large coverage loss. They leave open possible benefits on
harder tasks, smaller models, long multi-turn work, or decisions that depend on
recent specialist knowledge.

\paragraph{What is distillation worth?}
A static artifact has engineering advantages that E4's accuracy comparison
does not capture. It can be read, carried, and loaded cheaply many times. It
can also provide a stable map of a person's thinking instead of fetching fresh
passages for every question. Looking up passages as needed can provide fresher
material and more surrounding context, but it requires infrastructure and can
miss the source, as it does on four E4 questions. Our evidence does not tell us
which trade-off users prefer or whether the map is faithful. Answering those
questions requires human source audits and a larger retrieval comparison,
including stronger semantic-vector retrievers and combinations of keyword and
semantic retrieval.

The quotation result has a similar limit. A matching span can catch some
invented wording, but overlapping text does not establish who said it or
whether the claim follows. Systems for citation-supported generation treat
citation support, citation completeness, and factual precision as separate
measures~\citep{menick2022verified, gao2023rarr, gao2023alce,
min2023factscore}. Future versions of \mimeo{} should measure them separately
too.

\paragraph{The self-critique loop optimizes its own checklist.}
The authoring model scores its own draft under the same rubric it uses for
revision, then that score selects the best draft. The increase shows that the
loop optimizes its checklist. It does not show that a human expert would find
the file more faithful. The clustering ablation exposes this limit: cutting the
item count by half or more through clustering leaves the checklist score
unchanged. An end-to-end evaluation needs raters who know the field. They must
inspect source identity, speaker attribution, whether each claim follows from
its source, what dissenting material was left out, and whether the system
turned a recurring theme into a rule that is too broad.

\begin{figure}[t]
  \centering
  \includegraphics[width=\textwidth]{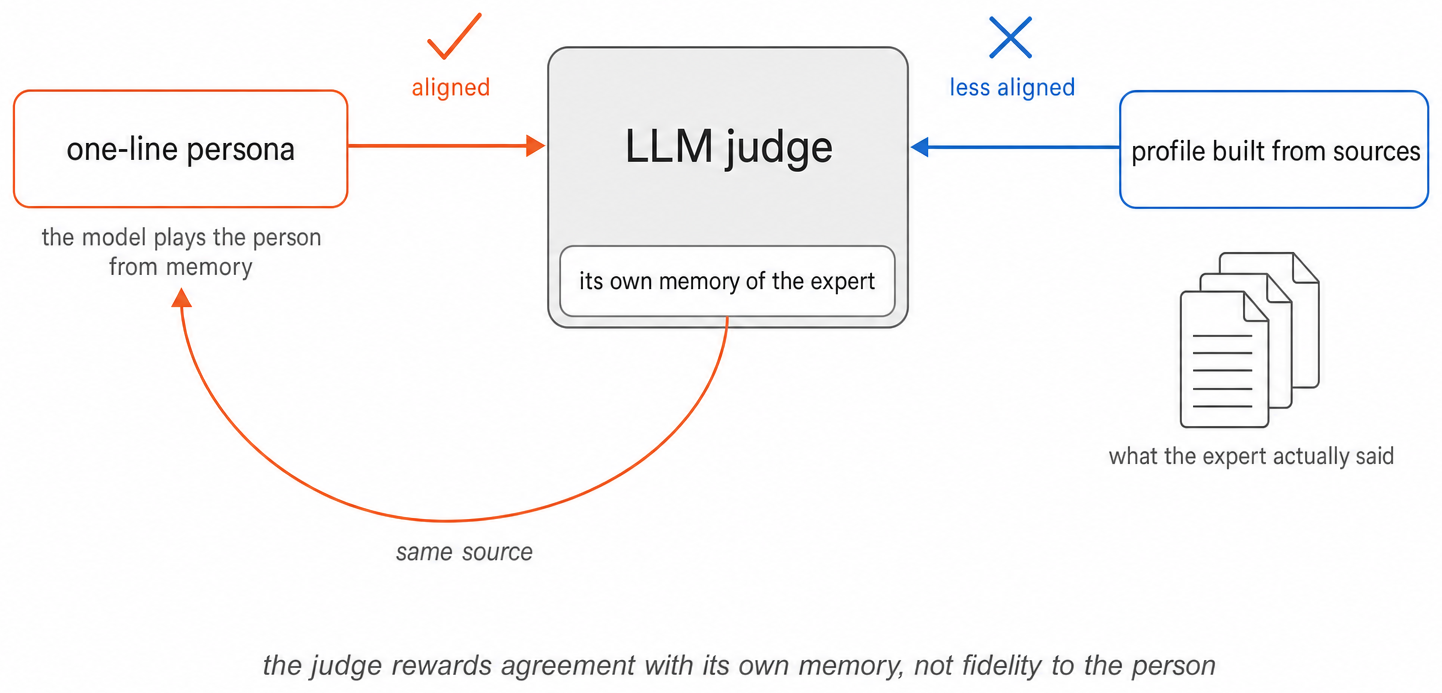}
  \caption{A possible failure mode when a judge evaluates a persona from its
  own memory. The standard prompt asks a judge to compare an answer with what
  it already knows about the named expert. An answer drawn from the same kind
  of memory may therefore look more faithful than one carrying less familiar
  corpus detail. Two of our four judges significantly prefer the
  stereotype-based answers in the pooled test; two do not detect an effect.
  The left box stands for both ungrounded conditions: the pooled test covers
  112 pairs, 56 against the one-line persona and 56 against the from-memory
  profile. The diagram states a hypothesis consistent with that split. It is
  not a demonstrated cause, and the panel does not establish one.}
  \label{fig:circularity}
\end{figure}

\paragraph{Whether an answer sounds like the expert depends on who is asked.}
Table~\ref{tab:judges} in the main text reports the panel analyzed here. The
original pairwise prompt asks which answer better reflects a named person's
``publicly documented principles, heuristics, and way of reasoning,'' based on
what the judge already knows. Four models from four developers scored the same
112 pairs. Their pooled \mimeo{} shares are .37, .51, .41, and .47. The first
and third are significantly below the .5 no-preference mark; the other two do
not detect an effect. No judge significantly prefers \mimeo{}. This result
fails to replicate across the fixed panel. It is not a sign reversal or an
estimate over LLM judges in general. Figure~\ref{fig:circularity} sketches one
mechanism that would produce this split.

The differences between judges are larger than those produced when we shuffle
the judge labels within each pair ($p=9.5\times10^{-4}$). This remains true when
we shuffle once per task or task-by-expert cluster ($p=.0032$ and $.0016$).
ICC(2,1) is .50: stable differences among answer pairs account for 49.5\% of
the variation, judge-wide shifts for 2.0\%, and residual disagreement specific to a pair and judge for 48.5\%. Agreement between judges on individual
alignment judgments is Krippendorff's $\alpha=.18$ in E2 and $.58$ in E5.
JudgeBench also finds that judging preference and judging
correctness are different things~\citep{tan2025judgebench}. A panel can expose
this instability, but it cannot say from outside what Karpathy or Sutton would
do.

Exploratorily, the two judges that show the pooled preference for the
stereotype are also the two most consistent with themselves when answer order
is swapped \emph{on the alignment question}, in both E2 and E5. The pattern
does not carry over to judged quality. There, \texttt{gpt-5.6-terra-pro} is the
most order-consistent judge on E2 and ties \texttt{claude-opus-5} on E5. This
pattern across two judges and two experiments is worth testing, but it does not
yet explain judge behavior.

% Generated by experiments/common/make_tables.py -- do not edit by hand.
\begin{table}[t]
  \centering\small
  \setlength{\tabcolsep}{5pt}
  \begin{tabular}{lccccl}
    \toprule
    Judge & E2 quality & E2 alignment & E5 quality & E5 alignment & Alignment state \\
    \midrule
    \texttt{grok-4.6} (xAI) & 0.70 & 0.72 & 0.60 & 0.93 & below parity \\
    \texttt{claude-opus-5} (Anthropic) & 0.54 & 0.58 & 0.75 & 0.88 & below parity \\
    \texttt{gpt-5.6-terra-pro} (OpenAI) & 0.72 & 0.54 & 0.75 & 0.79 & no effect \\
    \texttt{deepseek-v4-pro} (DeepSeek) & 0.32 & 0.42 & 0.35 & 0.71 & no effect \\
    \bottomrule
  \end{tabular}
  \caption{\textbf{Position consistency, every judge.} The fraction of pairs whose two judgments agree once the answers are swapped between positions. A judge who ignores order entirely scores $1.0$; what counts as chance agreement depends on how often that judge uses wins, losses, and ties. Rows are ordered by E2 alignment consistency. The last column reports each judge's registered state on the pooled alignment test.}
  \label{tab:consistency}
\end{table}

Position consistency cannot be compared directly with .5 when a judge can call
a tie. The chance rate depends on how often that judge uses wins, losses, and
ties. We therefore report the plain fraction of swapped pairs that agree for
each judge in Table~\ref{tab:consistency}. A future study should measure order
effects with randomized orders, calibrate candidate judges on examples with
known answers, and include human raters who know the experts' work.

\paragraph{Why measures with an outside reference behave better.}
Coverage of a fixed issue list, grading against a fixed answer key, quotation
matching, and identification with shuffled names all have a reference beyond a
judge's impression. These measures have their own limits, but changing the
grader rarely changes the overall ordering. Judged advice quality and expert
alignment have no outside key. The first gives a quality penalty under two
judges and none under two; the second gives the split above. We therefore leave
the quality penalty out of the paper's conclusions and use alignment only as a
result about measurement.

\section{Ethics and Limitations in Full}
\label{sec:ethics}

\paragraph{Profiles of real people.}
\mimeo{} builds artifacts about identifiable people, most of them living.
Making material public does not mean consenting to an AI-written profile. A
person's web record is incomplete and dated. It also depends on which languages,
venues, and publishers a search can reach. Distillation may preserve a position
the person has since abandoned, turn a conditional remark into a rule, or make a
model-written summary seem more authoritative than its sources. The named person
did not write, approve, or endorse the artifact.

The files use the third person, and the gallery labels them as generated. They
do not claim to be the person. Matching quotations to cached source text reduces
one risk but does not remove it. A close text match does not establish who spoke,
and the system has already credited a quotation to the wrong source. A generated
file should never be presented as the person's own words or as evidence of their
endorsement.

Our measurements show one way an ungrounded profile can misrepresent someone.
On the 20 single-record E4 questions
(Appendix~\ref{sec:regimes-knowledge}), the one-line persona and the from-memory
profile attribute a position other than the documented one on 1--4 of 20 answers
under every grader. The no-skill condition and \mimeo{} do so zero times. A
persona file based on a model's impression of a person therefore converts an
honest refusal into a confident wrong answer. This result supports grounding, but
does not establish that grounded artifacts are safe. \mimeo{} still carries
every risk above, and the single-record probes are a narrow test. The graded category also
covers an answer that names a different plausible position rather than only one that
contradicts the record, and 20 probes cannot support a probe-level test of the
difference.

The released toolkit and profiles provide no process for a named person to
review, correct, or remove a profile. This is an unresolved deployment defect,
not a minor feature to add later. Any hosted service should name a contact,
publish a response time, show a visible correction history, and remove profiles
promptly on request. It should also keep synthetic portraits off unless the
person consents. Portrait generation is already off by default in the CLI.

\paragraph{Eliciting private expertise.}
Interviewing or shadowing an expert would create different risks from compiling
public sources. The expert's consent would not extend to patients, colleagues,
or organizations whose information appears during that work. In medicine or
science, a system would need data minimization, meaning that it collects only
the data it needs. It would also need domain-specific oversight, separate consent
from affected people, and a review process that lets the expert correct or
delete both the source records and the derived skill. Raw observations should
not become reusable agent context by default.

\paragraph{Copyright and source handling.}
The pipeline caches excerpts, extracted pages, captions, and optional audio
transcriptions. Public access does not grant redistribution rights. Users remain
responsible for source licenses, quotation limits, database terms, and takedown
requests. Released artifacts should use citations and short quotations instead
of republishing cached source bodies. The toolkit and profiles released with
this paper include neither the experiment runs nor the cached source bodies. A
production service also needs retention limits and a way to remove cached
content when its source is withdrawn.

\paragraph{Discovery bias and identity errors.}
The strongest corpora come from English-speaking, prolific, web-visible
experts. Coverage is uneven when work sits behind paywalls, appears in audio
that cannot be fetched, comes from a thin archive, or is published in another
language. Deduplicating URLs does not detect mirrors or syndicated copies.
Record counts can therefore overstate the number of independent sources that
support an item.

The least visible person in the original gallery exposes a more serious
failure. Roughly a third of his retrieved records are directory pages, and one
record belongs to a newspaper writer with the same name. Identity
disambiguation identifies the target before discovery, but the pipeline does
not then check each retrieved record against that identity. We exclude this
profile from the experiments. Before publishing a profile, a deployable system
should verify that every record belongs to the right person, detect copied
sources, and show the source list to that person.

\paragraph{Prompt injection and network safety.}
Fetched pages are untrusted input. The implementation wraps them in data markers
and rejects URLs that carry credentials, point into private networks, or return
oversized responses. Those markers do not form a security boundary: a malicious
page can still try to influence extraction or authoring. Because we did not run
an adversarial prompt-injection benchmark, the system should run without secrets
or write access to sensitive repositories. Future work should test these attacks
at each stage and record which source produced a contaminated item.

\paragraph{Evaluation scope.}
The behavioral evidence comes from one commercial agent model and one
coding-agent harness. The model keeps changing. We record model names centrally,
but a vendor alias does not lock the underlying model weights. The committed
outputs support offline analysis, while new runs may give different results
after providers update their models. Any future release should therefore include
execution dates, CLI versions, prompts, and the model metadata returned by the
provider.

The paper's author wrote the original tasks and answer keys. Three new
generations for each keyed combination show how much answers vary. However, each
of the four \mimeo{} profiles and four from-memory profiles is built only once.
The design combines every task with every expert, but includes only four keyed
tasks and four experts. Judging, rather than generation, limits the size of this
grid: each added task or expert increases the number of answers, and those
answers then need repeated scores from four judges and two graders. We summarize the main contrasts at the task
level so that repeated answers within one task do not count as independent
evidence. This is how we avoid pseudoreplication, or counting repeated answers
as independent. Mixed-effects analysis faces the same broader problem with
repeated items~\citep{baayen2008mixed}, but that paper does not prescribe our method. The
resulting intervals are still estimates over a very small set of tasks.

The engineering coverage test and the E6 application test are both at their
ceiling. They rule out a large loss, but cannot measure a gain of the same size.
E4 is mostly a test of quotation recall from the same corpus carried by the
treatment. Its BM25 arm provides a retrieval control, but has only 40 questions
and BM25 is not a strong retriever. E4b has 18 questions, two graders, no
retrieval arm, and no fixed record written before the outcome was known. E5b
tests one synthetic evidence packet; packets of other lengths or document types
may produce different results.

No human rater in this study is an expert on all four people. LLM graders
working from corpus keys agree well with one another. They still cannot
determine whether the distilled principles capture each person's judgment in
context. Evaluating that claim requires human source audits and scenario ratings
from people who know the work. We did not recruit such raters, so we do not
describe the artifacts as digital twins, clones, or verified expert reasoning.

\paragraph{Compute.}
The study makes API calls in three places: the pipeline re-runs, the 732 agent runs,
and the judging and grading passes. Judging and grading use most of the compute.
Four judges repeat the pairwise and identification judgments and the main E4
and E6 grades; two graders score E2 coverage and E4b. Generating an artifact
uses far less compute than judging it repeatedly. Several primary-judge calls
failed and were retried before they returned parseable results. The telemetry
for successful outputs omits those attempts, so the recorded totals understate
actual usage. The per-run pipeline telemetry in
Appendix~\ref{sec:intrinsic-reruns} covers model calls only. It excludes local
compute and the optional transcription and image stages.

\section{Evaluation Task Suite}
\label{app:tasks}

Appendix~\ref{sec:extrinsic} uses eight tasks that call for judgment. Each prompt
contains all its own materials, and none names an expert. Four
tasks have a planted issue list that agents and judges never see. They are tasks
1, 3, 5, and 7 below: code review, debugging strategy, experiment design, and
performance investigation. Their lists contain 8, 3, 6, and 4 issues,
respectively, giving the 21 issues per expert reported in
Appendix~\ref{sec:extrinsic-design}. The other four tasks also contain planted
flaws. Those counts describe the task design, however, and do not come from a
list used for scoring. The scored list is \emph{closed}: each keyed task ships
with a \texttt{key.json} that names exactly its planted issues. The grader must
return one verdict for every numbered issue. The key, rather than the grader,
therefore determines how many issues count toward coverage. The tasks and their
planted issues follow.

\begin{enumerate}[leftmargin=1.5em]
  \item \textbf{Code review} of a $\sim$120-line PyTorch training script with eight
  planted issues (validation-set leakage through normalization statistics, transform
  aliasing that silently strips augmentation, graph-retaining loss accumulation,
  missing \texttt{no\_grad} in validation, a per-batch scheduler step fighting a
  manual warmup, a zero-LR warmup off-by-one, unscaled fp16, incomplete seeding).
  \item \textbf{Architecture critique} of an over-engineered retrieval-augmented
  generation (RAG) design for document QA in a 50-user internal tool (seven
  microservices, three databases, an agent swarm), with ten planted flaws and a
  request for a counter-proposal.
  \item \textbf{Debugging strategy} for a training run whose loss plateaus, spikes,
  and diverges from validation after learning-rate (LR) decay; three planted root
  causes, four red herrings, and a baseline that mixes two changes.
  \item \textbf{Refactoring plan} for an organically grown 20k-line research codebase
  (global config, five divergent training loops, unreproducible experiments), with
  constraints that penalize full rewrites and research freezes.
  \item \textbf{Experiment design} to establish or refute a claimed optimizer
  improvement whose evidence has six planted weaknesses (a tuned method compared
  against a default one, hyperparameters tuned on the test set, a single seed, a
  single small workload with generalization asserted rather than tested, hidden
  costs left out of the headline claim, and a ``fewer epochs to 90\%'' schedule
  artifact).
  \item \textbf{API design review} of a proposed model-evaluation library with seven
  planted families of code smells (hidden global state, excessive use of boolean flags,
  silently swallowed kwargs, eager side effects, untyped NaN failure results,
  runtime input sniffing, and a cache keyed on object identity).
  \item \textbf{Performance investigation} of a 40\% training-throughput drop after a
  cluster migration; four planted causes with distinct signatures, four red herrings,
  and a smoke test that changes two things at once.
  \item \textbf{Incident postmortem} for a silent feature-pipeline schema drift that
  served garbage predictions for six hours; three planted root causes, four missed
  detection opportunities, and details designed to invite misplaced blame that
  the response should resist.
\end{enumerate}

\section{Open-Ended Task Suite (E5)}
\label{app:e5tasks}

Appendix~\ref{sec:regimes-openended} uses six short advisory prompts. Unlike the
Appendix~\ref{app:tasks} suite, they include no supporting materials. Each
scenario demands firm choices. With no supporting material, little else in the
context competes with the always-on profile. None names an expert. The six
prompts are:

\begin{enumerate}[leftmargin=1.5em]
  \item \textbf{Research agenda} for a new three-person applied-AI lab
  (twelve months, eight GPUs): pick two or three directions, name what you
  refuse to work on, and define month-twelve success.
  \item \textbf{Startup direction} for a six-person model-serving startup with
  flat revenue: commit to one direction, no menu of options.
  \item \textbf{Course design}: a twelve-week graduate seminar on doing
  excellent applied AI work, with a mandatory semester project.
  \item \textbf{Hiring loop} for a senior ML engineer: stages, signals, and one
  standard practice to drop.
  \item \textbf{Open problems}: the three most important open problems in AI,
  ranked, with credible lines of attack.
  \item \textbf{PhD advice}: how a first-year student should pick a thesis
  topic and spend their first two years.
\end{enumerate}

\section{Paired Materials and Application Suites}
\label{app:newtasks}

\paragraph{E5b materials.}
E5b repeats the six E5 requests word for word, then adds two blocks: a
task-specific packet of 1{,}748--1{,}970 characters and a shared organizational
record of 5{,}105 characters. Together, these blocks add
6{,}853--7{,}075 characters. They contain constraints, measurements, and
competing considerations, but no expert name or quotation. Every E5/E5b pair
keeps the task, condition, and expert fixed; only the packet changes.

\paragraph{E6 application probes.}
The 16 scenarios include four for each expert, with each scenario based on a
different corpus principle. None of these principles is a quotation, and the
prompts never name the expert. Before running E6, we wrote
\path{experiments/e6_application/ANALYSIS_PLAN.md} and \texttt{probes.json};
every run records their joint SHA-256 digest \texttt{a02708e...c9062}. This is a
plan fixed locally in advance, not a registration with an external timestamp.
The key specifies both the expected decision and the distinctive reasoning
behind it. Four graders label each answer as correct, partial, or contrary to
the key.

\section{Persona Conditions}
\label{app:conditions}

The sandbox working directory receives all persona content as
\texttt{CLAUDE.md}. The agent harness reads this file at the start of a session,
and \mimeo{}'s \texttt{AGENTS.md} output is written for this deployment. The no
skill condition writes no file.

The on-demand arm (Appendix~\ref{sec:extrinsic-ondemand}) is the exception. It
does not write \texttt{CLAUDE.md}. Instead, it installs the bundle at
\texttt{.claude/skills/$\langle$expert$\rangle$/SKILL.md} with its
\texttt{references/} directory. The agent can find the skill by its description,
but loads it and the supporting reference files only if it chooses to do so.
These runs record the full event stream rather than only the final answer. Whether the
agent loads the skill is an outcome of this arm rather than part of its setup.
Because the summary output reports turn counts but not tool calls, we count a
run as loading the skill when its event stream contains a skill invocation that
names the expert.

% \mbox{} closes the run-in paragraph head so the listing frame starts on its
% own line instead of being drawn through the heading.
\paragraph{One-line persona (example).}\mbox{}
\begin{lstlisting}
You are Andrej Karpathy -- deep learning researcher and educator,
former Director of AI at Tesla, founding member of OpenAI. Approach
every task the way Andrej Karpathy would: reason, prioritize, and
communicate as they do.
\end{lstlisting}

\paragraph{From-memory profile.} The model that authors the \mimeo{} pipeline
(\texttt{google/gemini-3.6-flash}) also wrote this profile, but had no sources in
front of it. It used the same structural template as \mimeo{}'s
\texttt{AGENTS.md} author prompt and matched the length of the \mimeo{}
artifacts in the study: 2{,}382--2{,}918 words, compared with \mimeo{}'s
2{,}513--3{,}251, with means within 4.5\%. This comparison holds the authoring
model, broad format, and approximate length fixed. It does not isolate one
factor. Grounding in a corpus also changes the claims, quotations, source
identifiers, and editing process. E4's BM25 condition is the direct control for
access to the same cached corpus.

\section{Statistical Procedures}
\label{app:stats}

The implementations are in \texttt{robustness.py} and
\texttt{make\_tables.py}, under \texttt{experiments/common/}. They derive every
statistic reported below from the recorded raw judgments.

\paragraph{Pair scores.} We judge every pairwise comparison twice, swapping the
positions of the two answers the second time. Each judgment gives \mimeo{} a
score of $1$ for a win, $0$ for a loss, or $0.5$ for a declared tie. We average
the two judgments into one \emph{pair score}, so its possible values are
$\{0, 0.25, 0.5, 0.75, 1\}$. For example, one win and one tie gives $0.75$ and
counts as favoring \mimeo{}. The reported \emph{share} is the mean pair score
across cells; $0.5$ means no preference. \emph{Position consistency} is the
fraction of cells for which the two judgments agree after we account for the
swapped answer order. A judge who ignores order entirely scores $1.0$. The
level of chance agreement depends on how often that judge gives wins, losses,
and ties.

\paragraph{Tests.} For pairwise comparisons, the two-sided exact sign test uses
only pairs with a decided outcome. We report the Wilcoxon signed-rank test on
pair scores alongside it because that test also captures the size of split
decisions. Paired, question-by-question comparisons in E4 use the exact McNemar
test~\citep{mcnemar1947}. McNemar looks only at questions on which the two
conditions disagree. Binary proportions use Wilson
intervals~\citep{wilson1927}. The pair-score intervals in Figure~\ref{fig:extrinsic}b are Wilson-style
with fractional successes. They are descriptive, do not
match a binomial sampling model, and do not support a null claim. For the main
repeated-coverage and identification results, we aggregate observations within
tasks or resample task means, as described below.

\paragraph{Blind identification.}
The original question offers four names plus ``none of them,'' always in the
same order. Its very low primary-judge scores are entangled with a 75--81\%
abstention rate. The fixed order also makes it impossible to separate a
preference for one list position from the prominence of the expert in that
position. We retain these data as a record of the original design, but they are
not the main identification result.

The replacement question forces a choice among four names and shuffles their
order for each answer. A seed derived from a SHA-256 hash keeps every shuffle
fixed and reproducible. All four judges see the same order, and chance is
1/4. The repeated E2 analysis has 48 answers per persona condition: four
tasks, four experts, and three generations. E5 and E5b have 24 each. We obtain
accuracy intervals by resampling task means. For the paired E5/E5b test, we
average experts, conditions, and all four judges within each of the six tasks,
then enumerate all $2^6$ ways to flip the task-level signs. This analysis treats
tasks, rather than repeated judgments, as the independent units.

\paragraph{Multiplicity.} Within each experiment, the six pairwise comparisons
(two judged axes $\times$ three baselines) form one family of related tests. We
adjust them with the Holm--Bonferroni step-down
procedure~\citep{holm1979}. Holm controls the chance of at least one false
positive within that family. Both inference tables report raw and adjusted
$p$-values. We do not correct across experiments or measures. We treat the
primary judge's pooled directional test in Appendix~\ref{app:discussion} as a
discovery result, not a confirmatory test of a hypothesis stated in advance.
The later registration classifies the third and fourth judges separately; it
does not turn the primary result into a hypothesis stated in advance.

\paragraph{Equivalence and power.}
For coverage, we use TOST~\citep{lakens2017tost} to test against a margin of
$0.10$ of the key. TOST is an equivalence test: it asks whether the difference
is small, rather than whether it differs from zero. A tenth corresponds to
0.3--0.8 whole issues across the four task keys, or roughly one half-credit
partial issue in the median task. In the repeated analysis, we average the four
experts and three generations within each task before testing. Each comparison
therefore has four task-level units. TOST is symmetric, but the practical
interpretation is not. The margin rules out a loss large enough to change what
we would advise deploying, while the 95--97\% control ceiling makes an equally
large gain impossible to observe. We therefore call this result
ceiling-limited non-inferiority: evidence that the tested condition is not
worse by the chosen margin, limited by scores already near the maximum. It is
not symmetric evidence of no effect. We also report a conditional pair-level
minimum detectable effect, which is the smallest pair-level effect the original
analysis could reliably detect at the observed tie rate. It does not cover the
whole design and shows that these analyses could miss modest preferences.

\paragraph{Judge panel.} Four models from four different developers served as
judges. Each independently evaluated every answer pair, identification
response, and E4 answer:
\texttt{claude-opus-5} (Anthropic) through the Claude Code CLI, and
\texttt{gpt-5.6-terra-pro} (OpenAI), \texttt{grok-4.6} (xAI), and
\texttt{deepseek-v4-pro} (DeepSeek) through OpenRouter. They receive identical
prompts and answers. The three OpenRouter calls use temperature 0; the Claude
Code CLI does not expose a temperature setting in this harness. No two judges
share a developer, and the primary judge uses a different inference route from
the three replicates. We report all four judges. A result counts as established
only if it holds for every one of them (Table~\ref{tab:judges}).

Google is deliberately absent from the panel. \texttt{gemini-3.6-flash} wrote
both the \mimeo{} artifacts and the from-memory profiles, so a Gemini judge
would score its own writing. We could not distinguish a preference for that
writing from the ordering we are testing.

Two judges can show that two models disagree, but they cannot distinguish ``one
judge is an outlier'' from ``this measure is unstable.'' That distinction
requires a panel. We report three further quantities. First, because the judges
score the \emph{same} pairs, their shares are paired rather than independent
estimates. We test whether judges differ by shuffling judge labels within each
pair for $20{,}000$ draws. Under the null assumption, a judgment is a property
of the pair rather than the judge, so judge identity does not matter. We also
apply one judge-label permutation to every observation within a task or
task-by-expert cluster. Both tests show that judges differ ($p=.0032$ and
$.0016$). Every direct comparison between judges is paired for the same reason.
Overlapping intervals do not show that two judges agree. The six paired mean differences run from $-.143$ to $+.107$, and three of
their bootstrap intervals exclude zero: \texttt{opus-5} against
\texttt{gpt-5.6} at $-.143$ $[-.221, -.065]$, \texttt{gpt-5.6} against
\texttt{grok-4.6} at $+.107$ $[.040, .174]$, and \texttt{ds-v4} against
\texttt{opus-5} at $+.103$ $[.025, .179]$.

Second, we report panel agreement with Krippendorff's
$\alpha$~\citep{krippendorff2004}. Krippendorff's alpha measures agreement while
allowing both nominal labels, whose categories have no order, and ordinal
scores, whose values do have an order. We retain the pairwise Cohen's $\kappa$
matrix so the two-judge figures remain traceable. We use nominal $\alpha$ for
verdicts and grades and ordinal $\alpha$ for pair scores. The six nominal
values are .20 for E2 advice quality, .18 for E2 alignment, .26 for E2
identification, .21 for E5 advice quality, .58 for E5 alignment, and .81 for
E5 identification. The ordinal $\alpha$ for the pooled pair scores is .49. Fleiss' $\kappa$ provides a nominal
cross-check. It differs from $\alpha$ by at most .002 across the six panel
instruments. The registered J1 outlier diagnostic does not identify J1 as an
agreement outlier: its mean pairwise $\kappa$ is .336, compared with .281 for
pairs that exclude it.

Third, we report ICC(2,1) in the Shrout--Fleiss
sense~\citep{shrout1979icc}: two-way random effects, absolute agreement, single
rater. This intraclass correlation coefficient is the share of total variation
attributable to stable differences among answer pairs. The remainder combines
stable judge differences with residual pair-by-judge disagreement. Thus,
$1-\mathrm{ICC}$ is not simply ``variation caused by the judge.'' ICC helps
show whether one judge's persona score can stand in for the panel's measurement.
Position consistency, reported for each judge in
Table~\ref{tab:consistency}, is computed by judged axis and experiment. We measure
agreement between judges for each judgment with Cohen's
$\kappa$~\citep{cohen1960kappa}; expected agreement uses both judges' marginal
label frequencies. For E4, we compute $\kappa$ on the correct-versus-not split
because that is the split reported in the paper.

Every saved pairwise and grading verdict parses, so each panel-wide analysis of
those measures includes every pair. Two of \texttt{deepseek-v4-pro}'s
identification records are the exception: the runner saved them with a corrupted
field name in place of \texttt{choice}. The judge's answer survives in both, and
both name the wrong expert, so reading them leaves every reported identification
number unchanged. Recovering an answer given as a name rather than a letter is
now part of the analysis, which also excludes a verdict it cannot resolve and
publishes that count. An earlier version of the analysis scored an unusable verdict as a wrong
identification, which would bias accuracy downward instead of showing a broken
record. The judge runners also did not retain failed parse attempts
that preceded a successful retry. The per-judge parse-failure counts promised
in the registration therefore cannot be reconstructed. We report both protocol
deviations instead of substituting the count of missing saved files.

\paragraph{Clustered contrasts.} The no-skill condition has no expert
attached, so its $32$ E2 pairs (and $24$ E5 pairs) reuse one control answer per
task across four experts. Counting every pair as independent would therefore
overstate the sample size. We collapse these comparisons to task means before
testing, leaving $8$ independent units in E2 and $6$ in E5. The direction and
rough magnitude do not change. No claim in the
paper rests on a clustered contrast alone.

The primary judge's pooled directional comparison in
Appendix~\ref{app:discussion} has the same structure: its $112$ pairs come from
$56$ distinct \mimeo{} answers, $14$ tasks, and $4$ experts. Treating them as
$112$ independent coin flips would overstate their independence. We therefore
report descriptive percentile intervals from $20{,}000$ resamples of whole
clusters. The share remains $0.37$: 95\% CI $[0.29,0.45]$ by task and
$[0.29,0.46]$ by task~$\times$ expert. These resamples come from the observed
distribution, not from a distribution built around the assumption of no
effect. They provide intervals, not null-test $p$-values.

The registered Outcome C plan proposed a confirmatory sign test after stacking
the third and fourth judges' ratings into $224$ observations. That test is
invalid because both judges score the same $112$ pairs, so we do not report it.
When we instead average the two ratings within each pair, the descriptive share
is $.44$, with a task-cluster interval of $[.38,.50]$. The individual
registered classifications remain the relevant result: the third judge is
below parity and the fourth is not. The ordering therefore does not replicate
across the panel.

\paragraph{Two graders on the closed key.} Coverage of the planted issues
carries the paper's equivalence claim. We therefore grade it twice, using
\texttt{claude-opus-5} and \texttt{gpt-5.6-terra-pro} with identical keys and
answers. We report agreement for each \emph{issue verdict}, rather than for each
answer, because issues are the units used to calculate coverage. On the
original pass, the graders agree exactly on 94.9\% of 273 verdicts
($\kappa{=}0.71$) and 95.6\% of found-versus-not decisions
($\kappa{=}0.75$). Across all three generations, they agree exactly on 95.1\%
of 819 verdicts and 96.2\% of the binary decisions. Both graders find
equivalence in all three repeated comparisons.

\section{The Familiarity Arm (E4b)}
\label{app:e4b}

\paragraph{Registration status.}
This arm is not preregistered. Our preregistration
covers only the four-judge replication and
contains no E4b outcome table. Code documents E4b's hypotheses and manipulation
check, but no record was fixed before the outcome was known. We therefore treat
the familiarity arm as exploratory. E6 has a plan and recorded digest fixed in
advance, but it is also not an external registration.

\paragraph{Isolation.} The arm never touches the registries shared by E2, E4,
and E5. This separation is necessary because the blind-identification prompt
builds its options from the expert registry. Adding an expert to that registry
would change the judge's prompt, invalidate 518 committed judgments, and change
the 1/4 chance rate. The arm therefore has its own registry and results
directories. It calls the shared code through keyword arguments while leaving
their defaults unchanged.

\paragraph{Corpus adequacy.} The model can fail questions about a less famous
expert because it knows little about that person, which is the effect we want
to measure, or because the pipeline failed to assemble a usable corpus. The
results alone cannot separate these explanations. We therefore fixed a screen
in advance. Each candidate needed at least 12 distilled items, at least 5 items
supported by two or more records, at least 8 matched quotations, and no more
than 20\% of sources from directory or profile pages. All three passed:
23--39 items, 11--20 items with more than one record, 22--31 matched quotations,
and 7--17\% directory pages.

\paragraph{Probes.} Each expert has six probes. Every probe is keyed to a corpus
item that appears in one record and contains a quotation matched to the cached
text. The builder requires each label to match exactly one item, so a question
cannot be linked to the wrong key without a warning. Each question also names a
\emph{distractor}: the specific wrong position that a model unfamiliar with the
person is most likely to substitute, usually a famous colleague's view. E4's
grader receives only a positive key. To mark an answer \emph{incorrect}, it must
decide that the answer contradicts the expert, which the key alone cannot
support if the grader also does not know the person. Naming the distractor in
advance makes this decision reproducible. Questions about Linda Griffith cover
only scientific and engineering positions. She has spoken publicly about her
own health, and we use nothing keyed to that subject.

\paragraph{Familiarity.} We use one measure for familiarity and another for the
outcome, so the familiarity check does not depend on the result. It uses free
recall rather than a keyed question. Instead of asking a judge, we score
responses by checking which words appear in the expert's own distilled corpus.
The headline measure is the rate at which the agent declines to answer because
this rate is least sensitive to how the corpus happens to be worded.

\section{Reproducibility}
\label{app:repro}

The public artifact released with this paper is the \mimeo{} repository, which
contains the toolkit and expert profiles.\footnote{\url{https://github.com/K-Dense-AI/mimeo}}
The experiment runs, judgments, prompts, keys, analysis scripts, and telemetry
patch are not part of that release. The 22 profiles analyzed in
Appendix~\ref{sec:intrinsic} are the profiles committed with the pinned toolkit
release.

\emph{Pins.} \texttt{experiments/common/models.py} records the model names used
by the active experiment runners. \texttt{google/gemini-3.6-flash} authors the
pipeline re-runs and from-memory profiles. \texttt{claude-sonnet-5} is the agent
under test, and \texttt{claude-opus-5} is the primary judge and grader.
\texttt{openai/gpt-5.6-terra-pro}, \texttt{x-ai/grok-4.6}, and
\texttt{deepseek/deepseek-v4-pro} are the replicate judges in
Appendix~\ref{app:discussion}. The historical gallery analyzed in E1 used
\texttt{google/gemini-3.1-pro-preview}, as recorded in
Appendix~\ref{sec:intrinsic}. \mimeo{} itself is vendored as a clone pinned to
one commit and patched only to add token and cost telemetry.

\emph{Resumability.} Every runner skips run IDs whose result file already
exists, so a killed long run can be relaunched. Each pipeline stage caches its
results under a fingerprint of its inputs, prompt text, and output schema.

\emph{Determinism.} Agent and judge calls are not deterministic, so re-running
the harness does not reproduce individual verdicts. However, every figure and
table in this study was regenerated deterministically from a fixed
\texttt{results/} tree, and the bootstrap used a fixed seed.

\section{Example Generated Skill (Excerpt)}
\label{app:example}

This excerpt opens the generated \texttt{SKILL.md} for Andrej Karpathy. It comes
from the released gallery; we omit reference files and attributions for space.

\noindent
\begin{lstlisting}
# Thinking like Andrej Karpathy

Andrej Karpathy approaches artificial intelligence and software
engineering through a "hacker's perspective" -- favoring code and
physical intuitions over dense mathematics. [...]

## Core principles

* Build from Scratch to Understand: To truly grasp complex systems,
  you must manually implement the core algorithms without relying
  on automated tools or copy-pasting [...]
* Keep the AI on a Leash: Because LLMs are fallible and possess
  "jagged intelligence," humans must verify their work in small,
  concrete chunks rather than trusting massive, fully autonomous
  outputs.

## Anti-patterns they push against

* Jumping to Full Autonomy: Trusting an AI to generate massive,
  unverified outputs (like a 10,000-line code diff) creates a
  massive verification bottleneck for the human.
* Trusting AI Demos: Believing a successful demo means the product
  is ready. Demos are works.any(); products are works.all().
\end{lstlisting}

\end{document}